%% file: main.tex
\documentclass[acmsmall]{acmart}
\usepackage{multirow}
\usepackage{booktabs}
\usepackage{adjustbox}
\usepackage{listings}
\usepackage{graphicx}
\usepackage{tabularx}
\usepackage{soul}
\usepackage{bm}
\usepackage{IEEEtrantools}
\usepackage{graphicx}
\usepackage{amsmath}
\usepackage{amsthm}
\usepackage{booktabs}
\usepackage{algorithm}
\usepackage{algorithmic}
\usepackage{multirow}
\usepackage{boldline}
\usepackage{hhline}
\usepackage{xcolor}
\usepackage{pifont}
\usepackage{makecell}
\usepackage{utfsym}
\usepackage[normalem]{ulem}
\useunder{\uline}{\ul}{}
\usepackage[inline]{enumitem}
\usepackage{enumitem}
\usepackage{IEEEtrantools}
\usepackage{stmaryrd}
\usepackage[edges]{forest}
\usepackage{soul}
\sethlcolor{red}
\usepackage{color,xcolor}
\usepackage{tikz}
\usepackage{utfsym}
\usepackage[normalem]{ulem}
\useunder{\uline}{\ul}{}
\usepackage[inline]{enumitem}
\usepackage{enumitem}
\usepackage{IEEEtrantools}
\usepackage{stmaryrd}
\usepackage{array}
\usepackage{tabularx}
\usepackage{tablefootnote} 
\definecolor{hidden-red}{RGB}{205, 44, 36}
\definecolor{hidden-blue}{RGB}{194,232,247}
\definecolor{hidden-orange}{RGB}{243,202,120}
\definecolor{hidden-green}{RGB}{34,139,34}
\definecolor{hidden-pink}{RGB}{255,245,247}
\definecolor{hidden-black}{RGB}{20,68,106}

\definecolor{hidden-red}{RGB}{205, 44, 36}
\definecolor{hidden-blue}{RGB}{194,232,247}
\definecolor{hidden-orange}{RGB}{243,202,120}
\definecolor{hidden-green}{RGB}{34,139,34}
\definecolor{hidden-pink}{RGB}{255,245,247}
\definecolor{hidden-black}{RGB}{20,68,106}

\definecolor{myGreen}{RGB}{127,210,85}
\definecolor{myOrange}{RGB}{242,154,66}
\definecolor{myYellow}{RGB}{247,223,65}
\definecolor{myRed}{RGB}{232,80,43}
\definecolor{myViolet}{RGB}{162,57,102}
\definecolor{myBlue}{HTML}{4686f3}
\definecolor{myYellowv2}{HTML}{E6C802}
\definecolor{myOrangev2}{HTML}{ED8E55}
\definecolor{MyGreenv2}{HTML}{009B55}
\definecolor{MyRedv2}{HTML}{c22f2f}
\definecolor{DarkRed}{RGB}{130,25,0}
\definecolor{PurpleRed}{RGB}{204,0,102}
\definecolor{DarkGreen}{RGB}{30,130,30}
\definecolor{DarkBlue}{RGB}{0,0,250}
\definecolor{DarkYellow}{RGB}{255,128,0}

\AtBeginDocument{%
  }

\setcopyright{acmlicensed}
\copyrightyear{2025}
\acmYear{2025}
\acmDOI{XXXXXXX.XXXXXXX}

\acmJournal{JACM}
\acmVolume{37}
\acmNumber{4}
\acmArticle{111}
\acmMonth{5}

\begin{document}

\title{A Survey of Slow Thinking-based Reasoning LLMs using Reinforced Learning and Inference-time Scaling Law}




\author{Qianjun Pan}
\authornote{Both authors contributed equally to this research.}
\author{Wenkai Ji}
\authornotemark[1]

\author{Yuyang Ding}

\author{Junsong Li}

\author{Shilian Chen}

\author{Junyi Wang}

\author{Jie Zhou}
\authornote{Corresponding authors.}
\email{jzhou@cs.ecnu.edu.cn}

\author{Qin Chen}

\author{Min Zhang}

\author{Yulan Wu}

\author{Liang He}
\affiliation{%
  \institution{School of Computer Science and Technology, East China Normal University}
  \city{Shanghai}
  \country{China}
}

\renewcommand{\shortauthors}{Pan et al.}

\begin{abstract} 
This survey explores recent advancements in reasoning large language models (LLMs) designed to mimic ``slow thinking"—a reasoning process inspired by human cognition, as described in Kahneman’s Thinking, Fast and Slow. These models, like OpenAI’s o1, focus on scaling computational resources dynamically during complex tasks, such as math reasoning, code generation, and agents. We present the development of reasoning LLMs and list their key technologies. By synthesizing over 160 studies, it charts a path toward LLMs that combine human-like deep thinking with scalable efficiency for reasoning. The review breaks down methods into three categories: (1) test-time scaling dynamically adjusts computation based on task complexity via search and sampling, dynamic verification; (2) reinforced learning refines decision-making through iterative improvement leveraging policy networks, reward models, and self-evolution strategies; and (3) slow-thinking frameworks (e.g., long CoT, hierarchical processes, hybrid thinking) that structure problem-solving with manageable steps. The survey highlights the challenges and further directions of this domain. Understanding and advancing the reasoning abilities of LLMs is crucial for unlocking their full potential in real-world applications, from scientific discovery to decision support systems.
\end{abstract}




\maketitle

\input{secs/01Introduction}

\input{secs/02Development}
\input{secs/03KeyTechnologies}
\input{secs/04InferenceTimeScalingLaw}
\input{secs/05ReinforcedLearningReasoning}
\input{secs/06SlowThinking}
\input{secs/08Challenges}

\section{Conclusions}
\label{Conclusions}
This survey provides a comprehensive exploration of the advancements, methodologies, and challenges in developing reasoning capabilities within LLMs. By tracing the evolution of prominent models and analyzing key technologies such as slow thinking, reinforcement learning, and knowledge distillation, we highlight the significant progress made in enhancing LLMs' ability to perform complex reasoning tasks. The synthesis of over 100 studies underscores the importance of categorizing research efforts into distinct paradigms—test-time scaling, reinforced learning, and slow thinking—each offering unique insights and trade-offs. 
Despite significant advancements, reasoning in LLMs is still far from achieving human-like robustness and flexibility. Key issues such as balancing fast and slow thinking, designing reliable reward mechanisms for reinforcement learning, ensuring interpretability, and integrating structured knowledge systems continue to pose formidable challenges.
This survey serves as both a foundation and a call to action for researchers and practitioners committed to advancing the frontier of reasoning in artificial intelligence.

\begin{acks}
This research is funded by the National Science and Technology Major Project (No. 2021ZD0114002), the National Nature Science Foundation of China (No. 62477010 and No. 62307028), the Natural Science Foundation of Shanghai (No. 23ZR1441800), Shanghai Science and Technology Innovation Action Plan (No. 24YF2710100 and No. 23YF1426100 ) and Shanghai Special Project to Promote High-quality Industrial Development (No. 2024-GZL-RGZN-02008).
\end{acks}

\bibliographystyle{ACM-Reference-Format}
\bibliography{main}


\end{document}

%% file: secs/01Introduction.tex
\section{Introduction}
The rapid advancement of LLMs like GPT-4 \cite{openai2023gpt4}, Deepseek \cite{liu2024deepseek}, LLaMA \cite{touvron2023llama,touvron2023llama2}, and Qwen \cite{bai2023qwen,yang2024qwen2} has revolutionized artificial intelligence, enabling breakthroughs in natural language understanding, code generation, and multi-modal reasoning. 
However, despite their impressive capabilities, traditional LLMs often struggle with tasks requiring deep, deliberate reasoning—capabilities typically associated with human ``System 2" cognition, as described by Kahneman in Thinking, Fast and Slow \cite{booch2021thinking}. While fast-thinking processes (``System 1") excel in quick, intuitive judgments, they falter in scenarios demanding sustained deliberation, such as solving complex mathematical problems or conducting nuanced ethical evaluations.

To address these limitations, researchers have turned to the concept of ``slow thinking" in AI—a paradigm inspired by human cognitive processes that emphasizes careful, step-by-step reasoning over rapid \cite{wei2022chain}, heuristic-driven responses \cite{min2024imitate,lawson2020comparing}. 
Recent innovations, such as OpenAI’s o1 \cite{openai_o1_system_card}, Deepseek R1 \cite{guo2025deepseek}, exemplify this shift by leveraging inference-time scaling laws to allocate computational resources during task execution dynamically. 
These slow-thinking-based reasoning systems are designed to tackle complex tasks by dynamically scaling computational resources during inference, a process referred to as inference-time scaling \cite{wu2024inference}. 
The integration of reinforced learning further enhances decision-making precision, enabling models to refine their strategies iteratively based on feedback and self-evaluation.
Such advancements have broadened the applicability of LLMs across domains like mathematics, visual reasoning, medical diagnosis, and multi-agent debates, where robust and reliable reasoning is paramount.

Recent advancements in the complex reasoning capabilities of LLMs have catalyzed substantial survey efforts within the academic community. These studies analyze the burgeoning field of reasoning LLMs from various perspectives: some concentrate on particular dimensions, such as the features and mechanisms of Long Chain-of-Thought \cite{chen2025reasoningerasurveylong,liu2023mathematical}; others emphasize core technical approaches, particularly positioning Reinforcement Learning (RL) as a primary driver for advancing model capabilities \cite{zeng2024scaling, xu2025largereasoningmodelssurvey}; certain works analyze model advancement via specific training paradigms, such as self-evolution \cite{he2025survey,tao2024survey}; while others offer comprehensive overviews of technical stages like post-training, including fine-tuning, RL, and test-time scaling \cite{kumar2025llmposttrainingdeepdive}. Notably, reference \cite{li2025system} aligns conceptually with the present work by also leveraging the dual-system theory (System 1/System 2) from cognitive science as an analytical framework.

Positioned within this context, the present survey provides a unique and focused perspective. Our primary contribution centers on the analysis of the cognitive goal of ``Slow Thinking" (analogous to System 2), distinctively identifying RL alongside the Inference-time Scaling Law as the principal, synergistic technical mechanisms for achieving this advanced reasoning capability. Although RL and test-time scaling are addressed individually or within wider scopes in other surveys \cite{zeng2024scaling, xu2025largereasoningmodelssurvey, li2025system}, this review uniquely synthesizes these two methodologies within the ``Slow Thinking" conceptual framework through systematic integration and in-depth analysis. 
Particularly, this survey aims to synthesize insights from over 160 studies to chart the evolution of slow-thinking-based reasoning LLMs. It categorizes current methodologies into three key areas: (1) test-time scaling, which adjusts computational effort based on task complexity; (2) reinforced learning, which enhances decision-making through reward models and self-improvement strategies; and (3) slow-thinking frameworks, which structure problem-solving using techniques such as long chain-of-thought (CoT) and hierarchical reasoning processes. 
Through this exploration, we aim to provide a comprehensive roadmap for future research in slow-thinking-based LLMs.

\begin{itemize}
    \item This survey provides a comprehensive overview of the evolution of reasoning Large Language Models (LLMs), detailing key technological advancements such as slow thinking mechanisms, reinforcement learning, and knowledge distillation. Additionally, we offer a comparative analysis of these models' defining characteristics and capabilities.
    \item By synthesizing insights from over 100 studies, this review organizes existing research into three primary categories: test-time scaling, reinforced learning, and slow thinking. A thorough analysis is conducted to evaluate and compare the performance and effectiveness of these approaches.
    \item Furthermore, we identify and discuss critical challenges and future directions for advancing reasoning in LLMs. These include achieving an optimal balance between fast and slow thinking, designing robust reward mechanisms, incorporating human-in-the-loop refinement processes, and addressing other open issues that remain pivotal for further progress.
\end{itemize}

%% file: secs/02Development.tex
\begin{figure}[t]
\centering
\includegraphics[width=1.0\columnwidth]{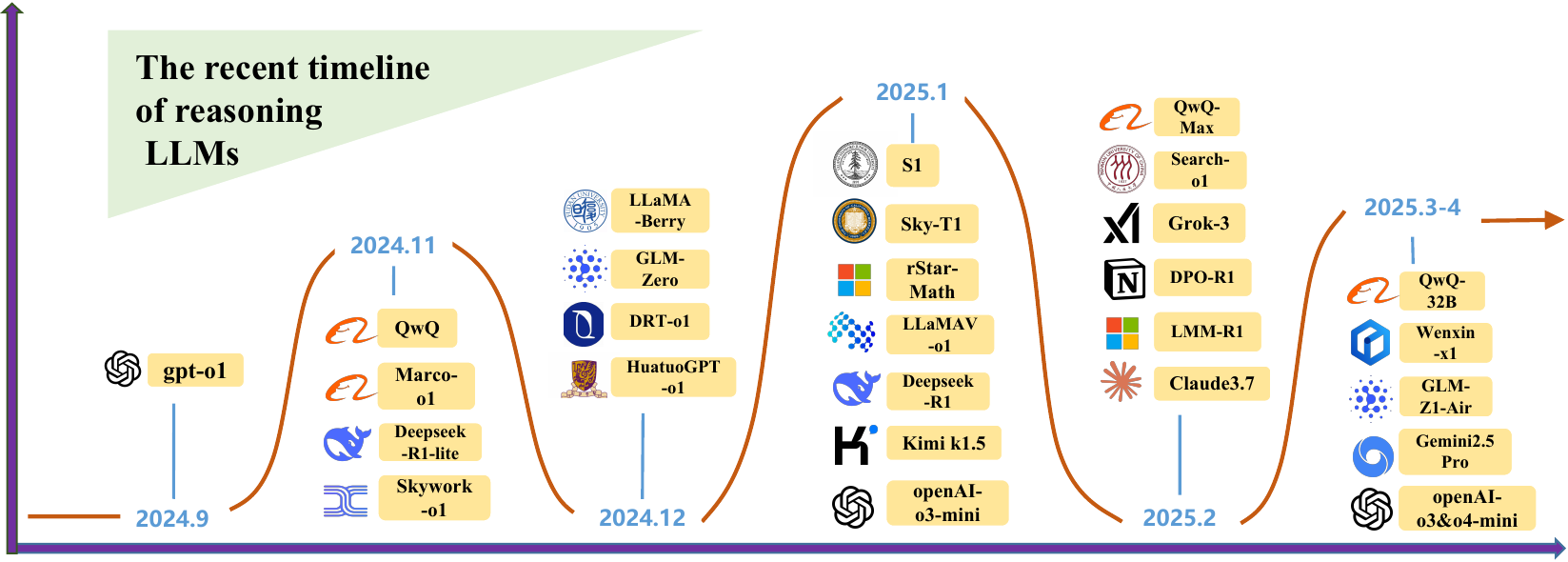}
\vspace{-5mm}
\caption{The timeline of main reasoning LLMs.}
\label{fig:timeline}
\vspace{-3mm}
\end{figure}

\section{The Development of Reasoning LLMs}
\label{sect:development}

Recently, o1-like models have emerged as a family of reasoning language models (RLMs) designed to replicate or extend the advanced reasoning, search, and alignment capabilities pioneered by OpenAI's o1 series \cite{zeng2024scaling,openai_o1_system_card} (Fig. \ref{fig: timeline}). These models typically share several core design principles. First, they incorporate reinforcement learning (RL) to optimize policy behaviors for complex reasoning tasks \cite{ElKishky2025CompetitivePW,guo2025deepseek,team2025kimi}. This often takes the form of Process Reward Models (PRMs) and Outcome Reward Models (ORMs), which respectively evaluate intermediate reasoning steps and final answers \cite{lightman2023let,guo2025deepseek,xu2024llava}. Second, they emphasize long COT or slow thinking paradigms, allowing the model to reason in multiple stages, verify partial solutions, and refine its outputs via techniques such as self-verification or guided search \cite{guo2025deepseek,kumar2025overthinkslowdownattacksreasoning,xu2024llava}. Third, search-based methods—for instance, beam search, Monte Carlo Tree Search (MCTS), or retrieval-augmented generation—serve as additional mechanisms to explore and validate candidate reasoning paths before finalizing an answer \cite{ElKishky2025CompetitivePW,zhao2411marco,zhang2024o1,zhong2024evaluation}.

Beyond these core techniques, many o1-like frameworks introduce multi-stage training pipelines that blend supervised fine-tuning (SFT) on human-curated CoT data with RL-driven policies for iterative refinement \cite{ElKishky2025CompetitivePW,guo2025deepseek,team2025kimi,kumar2025overthinkslowdownattacksreasoning}. This often includes cold-start or rule-based reward functions to bootstrap reasoning patterns, which are then iteratively replaced by more advanced reward models once the base model attains a certain level of proficiency \cite{guo2025deepseek,team2025kimi}. Active learning strategies, where high-uncertainty or error-prone samples are re-labeled or refined, further enhance data efficiency and model alignment \cite{lightman2023let,xu2024llava}. Some approaches also incorporate specialized modules—for example, retrieval-augmented generation (RAG) or verifiers—to manage external retrieval and mitigate hallucinations \cite{li2025search,xu2024llava,zhao2411marco,zhang2024o1}.

Collectively, o1-like models have demonstrated impressive performance on mathematical reasoning \cite{lightman2023let,team2025kimi,zhong2024evaluation} 
,
competitive programming \cite{ElKishky2025CompetitivePW,li2025search,zhong2024evaluation,zhang2024o1}, 
multilingual tasks \cite{openai_o1_system_card,xu2024llava}, multimodal reasoning \cite{huang2025vision}, and even domain-specific areas such as medical reasoning \cite{chen2024huatuogpt}, translation \cite{wang2024DRT-o1}. They underscore the growing trend toward systematic, step-by-step alignment of model outputs, balancing the benefits of open-ended thinking with robust error-checking. Despite these advances, open challenges remain in scaling RL pipelines without incurring ``reward hacking”, extending context windows to handle more complex queries, and mitigating potential biases in reward model design \cite{zeng2024scaling,guo2025deepseek,team2025kimi,xu2024llava}. Nevertheless, as evidenced by both commercial and open-source implementations \cite{guo2025deepseek,zhao2411marco,zhong2024evaluation,wang2024openrt}, o1-like models offer a promising blueprint for next-generation LLMs with stronger reasoning, verification, and alignment capabilities.


%% file: secs/03KeyTechnologies.tex
\input{tabs/evaluate}

\section{Key Technologies}
\label{Key Technologies}
In this section, we introduce the key technologies (e.g., slow thinking, reinforcement learning, reward model, knowledge distillation, search strategies and self-training) for reasoning LLMs. In Table \ref{table: evaluate}, we list the main reasoning LLMs and analyze their characteristics.

\subsection{Slow Thinking}
\paragraph{Theory of Slow Thinking}
In Daniel Kahneman's seminal work, Thinking, Fast and Slow \cite{daniel2017thinking,shleifer2012psychologists}, the differentiation between two modes of thought—System 1 and System 2—is central to understanding human cognition. System 1 represents fast, automatic, and intuitive thinking, while System 2 is characterized by slow, deliberate, and effortful reasoning. Slow thinking, as encapsulated by System 2, involves processes that require conscious attention, logical analysis, and mental effort. Unlike the rapid judgments of System 1, which are prone to cognitive biases, System 2 engages in careful evaluation and problem-solving, often correcting the errors introduced by the more impulsive System 1.

\footnotetext[6]{https://qwenlm.github.io/blog/qwq-max-preview}
\footnotetext[7]{https://www.anthropic.com/claude/sonnet}
\footnotetext[8]{https://x.ai/news/grok-3}
\footnotetext[9]{https://openai.com/index/learning-to-reason-with-llms/}
\footnotetext[10]{https://huggingface.co/Skywork/Skywork-o1-Open-Llama-3.1-8B}
\footnotetext[11]{https://api-docs.deepseek.com/news/news1120}
\footnotetext[12]{https://github.com/NovaSky-AI/SkyThought}
\footnotetext[13]{https://deepmind.google/technologies/gemini/} 
\footnotetext[14]{https://zhipuai.cn/devday}
\footnotetext[15]{https://console.bce.baidu.com/qianfan/ais/console/onlineTest/DeepSeek/ERNIE-X1-32K-Preview}
\footnotetext[16]{https://openai.com/index/introducing-o3-and-o4-mini/}

\paragraph{Slow Thinking for AI}

The principles of slow thinking are highly relevant to the development and application of LLMs. LLMs often emulate aspects of human cognition, potentially incorporating processes analogous to both System 1 and System 2 thinking. By integrating mechanisms that mimic System 2's deliberate reasoning, LLMs can better handle complex tasks demanding analytical depth and sustained focus.

One approach involves designing models combining fast pattern recognition (akin to System 1) with slow, reflective reasoning (akin to System 2). For instance, Booch et al. \cite{booch2021thinking} proposed frameworks where AI systems switch between rapid, pattern-based responses and slower, methodical evaluations for complex problems. Similarly, Qi et al. \cite{qi2024interactive} explored interactive continual learning methods using both fast and slow thinking to dynamically adapt to new information, thereby improving model performance over time.
Lin et al. \cite{lin2023swiftsage} demonstrated the value of slow thinking in generative agents like SwiftSage, which uses a dual-process architecture for complex interactive tasks. In this architecture, fast thinking provides immediate, heuristic outputs, while slow thinking ensures accuracy and coherence through iterative refinement and validation. This hybrid approach enables AI systems to balance efficiency and precision, overcoming limitations of purely reactive or purely deliberative models.

In summary, slow thinking (System 2) is crucial for enhancing the robustness and reliability of AI systems. Embedding mechanisms for deliberate reasoning allows LLMs to achieve greater sophistication, navigate nuanced scenarios, and deliver more accurate and considered responses. This integration not only reflects human cognitive strategies but also advances AI capabilities towards more adaptable and effective intelligence.

\subsection{Reinforcement Learning}
Reinforcement Learning (RL) \cite{kaelbling1996reinforcement,ernst2024introduction,arulkumaran2017deep} is a computational approach to learning whereby an agent interacts with an environment to maximize a cumulative reward. The agent learns an optimal policy, which dictates the actions to take in each state, through trial and error, guided by feedback in the form of rewards or penalties.
It is formalized as a Markov Decision Process $\langle \mathcal{S}, \mathcal{A}, P, r, \gamma \rangle$, where $\mathcal{S}$ represents the state space; $\mathcal{A}$ represents the action space; $P(s'|s, a)$ represents the transition probability function, defining the probability of transitioning to state $s'$ from state $s$ after taking action $a$; $r(s, a)$ represents the reward function, defining the immediate reward received after taking action $a$ in state $s$; $\gamma \in [0, 1]$ represents the discount factor, determining the importance of future rewards.

The agent learns a policy $\pi(a|s)$, which represents the probability of taking action $a$ in state $s$, to maximize the expected return:
\begin{equation}
    J(\pi) = \mathbb{E}_{\tau \sim \pi}\left[\sum_{t=0}^{\infty} \gamma^t r(s_t, a_t)\right]
\end{equation}
where $J(\pi)$ represents the expected return of policy $\pi$; $\mathbb{E}_{\tau \sim \pi}$ represents the expectation over trajectories $\tau$ sampled from policy $\pi$; $\tau = (s_0, a_0, s_1, a_1, \dots)$ represents a trajectory, a sequence of states and actions generated by following policy $\pi$; $t$ represents the time step; $s_t$ represents the state at time step $t$; $a_t$ represents the action taken at time step $t$; $\gamma$ represents the discount factor (as defined above); $r(s_t, a_t)$ represents the reward received at time step $t$ after taking action $a_t$ in state $s_t$.

\paragraph{Proximal Policy Optimization (PPO)}
PPO \citep{schulman2017proximal} is an on-policy RL algorithm that aims to improve a policy by taking small, safe steps. It optimizes a surrogate objective function that balances maximizing rewards and minimizing the deviation from the previous policy. The simplified objective function (assuming a single update after each exploration) is:
\begin{equation}
J_{PPO}(\theta) = \mathbb{E}_{q \sim P_{sft}(Q), o \sim \pi_{\theta_{old}}(O|q)} \left[ \frac{1}{|o|} \sum_{t=1}^{|o|} \frac{\pi_{\theta}(o_t|q, o_{<t})}{\pi_{\theta_{old}}(o_t|q, o_{<t})} A_t \right]
\end{equation}
where $q$ means the question from the supervised fine-tuning dataset ($P_{sft}(Q)$); $o$ means the output sampled from the old policy $\pi_{\theta_{old}}(O|q)$; $\theta$ means the current policy parameters; $\theta_{old}$ means the parameters of the previous policy; $o_t$ means the $t$-th token in the output sequence; $o_{<t}$ means the tokens preceding $o_t$ in the output sequence; $A_t$ means the advantage estimate at time step $t$.

\paragraph{Direct Preference Optimization (DPO)}
DPO \cite{rafailov2024direct,wang2024secrets} is an algorithm that directly optimizes the policy based on human preferences, bypassing the need for a reward model. It frames RL as a classification problem, learning to distinguish between preferred and dispreferred outputs. The objective function is:
\begin{equation}
\scriptsize
J_{DPO}(\theta) = \mathbb{E}_{q \sim P_{sft}(Q), o^+, o^- \sim \pi_{sft}(O|q)} \left[ \log \sigma \left( \beta \left( \frac{1}{|o^+|} \sum_{t=1}^{|o^+|} \log \frac{\pi_{\theta}(o^+_t|q, o^+_{<t})}{\pi_{ref}(o^+_t|q, o^+_{<t})} - \frac{1}{|o^-|} \sum_{t=1}^{|o^-|} \log \frac{\pi_{\theta}(o^-_t|q, o^-_{<t})}{\pi_{ref}(o^-_t|q, o^-_{<t})} \right) \right) \right]
\end{equation}
where $q$ means the question from the supervised fine-tuning dataset ($P_{sft}(Q)$), $o^+$ means the preferred output sampled from the SFT model $\pi_{sft}(O|q)$; $o^-$ means the dispreferred output sampled from the SFT model $\pi_{sft}(O|q)$; $\theta$ means the current policy parameters; $\pi_{ref}$ means the reference policy (typically the SFT model); $\beta$ means temperature parameter controlling the strength of preference; $\sigma$ means the sigmoid function.

\paragraph{Group Relative Policy Optimization (GRPO)}
GRPO \citep{shao2024deepseekmath} is a variant of PPO that simplifies the training process by foregoing the critic model. Instead, it estimates the baseline for advantage calculation from group scores, reducing computational resources. The objective function is:
\begin{equation}
\scriptsize
\begin{aligned}
    J_{GRPO}(\theta) = & \mathbb{E}_{q \sim P_{sft}(Q), \{o_i\}_{i=1}^G \sim \pi_{\theta_{old}}(O|q)} \\ 
    & \left[ \frac{1}{G} \sum_{i=1}^G \frac{1}{|o_i|} \sum_{t=1}^{|o_i|} \left( \frac{\pi_{\theta}(o_{i,t}|q, o_{i,<t})}{\pi_{\theta_{old}}(o_{i,t}|q, o_{i,<t})} \hat{A}_{i,t} - \beta \left( \frac{\pi_{ref}(o_{i,t}|q, o_{i,<t})}{\pi_{\theta}(o_{i,t}|q, o_{i,<t})} - \log \frac{\pi_{ref}(o_{i,t}|q, o_{i,<t})}{\pi_{\theta}(o_{i,t}|q, o_{i,<t})} - 1 \right) \right) \right]
\end{aligned}
\end{equation}
where $q$ means the question from the supervised fine-tuning dataset ($P_{sft}(Q)$); $\{o_i\}_{i=1}^G$ means a group of $G$ outputs sampled from the old policy $\pi_{\theta_{old}}(O|q)$; $\theta$ means the current policy parameters; $\theta_{old}$ means the parameters of the previous policy; $o_{i,t}$ means the $t$-th token in the $i$-th output sequence; $o_{i,<t}$ means the tokens preceding $o_{i,t}$ in the $i$-th output sequence; $\hat{A}_{i,t}$ means group-relative advantage estimate at time step $t$; $\pi_{ref}$ means the reference policy; and $\beta$ means a coefficient.


\subsection{Reward Model}

In the context of incentivizing reasoning capabilities in LLMs, reward modeling plays a pivotal role, as highlighted in the works of DeepSeek-R1 \cite{guo2025deepseek} and DeepSeekMath \cite{shao2024deepseekmath}. The reward model typically incorporates multiple components, such as outcome-based rewards, process-based rewards, rule-based rewards, and model-based rewards. Each component contributes uniquely to shaping the learning objectives.

\paragraph{Outcome Reward Model (ORM)}
The outcome reward model evaluates the correctness or accuracy of the final output generated by the model. This is often defined mathematically as:
\begin{equation}
R_{\text{outcome}} = \mathbb{I}(y_{\text{pred}} = y_{\text{true}})
\end{equation}
where $y_{\text{pred}}$ represents the predicted output, $y_{\text{true}}$ is the ground truth, and $\mathbb{I}(\cdot)$ is an indicator function that outputs 1 if the prediction matches the truth and 0 otherwise. Outcome rewards are critical for tasks requiring high precision, such as mathematical reasoning.

\paragraph{Process Reward Model (PRM)}
The process reward model \cite{uesato2022solving} focuses on rewarding intermediate steps or reasoning chains that lead to the final answer. It encourages the model to generate not only correct answers but also logically consistent reasoning paths. A common formulation for process rewards is:
\begin{equation}
R_{\text{process}} = \sum_{t=1}^{T} w_t \cdot \text{Quality}(s_t)
\end{equation}
where $s_t$ denotes the reasoning step at time $t$, $w_t$ is a weighting factor emphasizing important steps, and $\text{Quality}(\cdot)$ measures the quality of each step. This approach aligns with the methodology in DeepSeek-R1, which uses reinforcement learning to incentivize reasoning capabilities \cite{guo2025deepseek}.

\paragraph{Rule-based Reward}
Rule-based rewards incorporate predefined heuristics or constraints to guide the model's behavior. These include accuracy rewards and format rewards. Accuracy rewards ensure that the model adheres to task-specific correctness criteria, while format rewards enforce structural consistency, such as proper mathematical notation. For example:
\begin{equation}
R_{\text{rule}} = \alpha \cdot R_{\text{accuracy}} + \beta \cdot R_{\text{format}}
\end{equation}
where $\alpha$ and $\beta$ are hyperparameters balancing the contributions of accuracy and format rewards. Rule-based rewards are particularly useful in scenarios like mathematical reasoning, where strict adherence to symbolic rules is essential.

\paragraph{Model-based Reward}
Model-based rewards leverage auxiliary models to evaluate the quality of the generated outputs. These rewards are learned through supervised or reinforcement learning techniques. A typical formulation involves:
\begin{equation}
R_{\text{model}} = f_{\theta}(x, y_{\text{pred}})
\end{equation}
where $f_{\theta}$ is a reward function parameterized by $\theta$, $x$ is the input, and $y_{\text{pred}}$ is the predicted output. The reward function is trained to approximate human judgments or other gold-standard metrics. This approach is explored in the DeepSeek-R1 paper, which emphasizes the importance of learned rewards in enhancing reasoning capabilities.


\subsection{Knowledge Distillation}
Knowledge Distillation is a technique to transfer knowledge from a large, complex model (the \textit{teacher}) to a smaller, efficient model (the \textit{student}) \cite{xu2024survey}. The goal is to compress the teacher's expertise into the student while retaining performance. Two primary approaches exist: {Model Distillation} and {Data Distillation}.

\paragraph{Model Distillation}
Model Distillation directly transfers knowledge via the teacher's outputs.
The student learns by mimicking the teacher's softened probability distribution (soft labels) over classes, which contains richer information than hard labels (one-hot vectors). A temperature-scaled softmax is used to smooth the outputs, and the student's loss combines both the teacher's guidance and ground-truth labels 
\[
q_i = \frac{\exp(z_{t,i}/T)}{\sum_j \exp(z_{t,j}/T)}, \quad p_i = \frac{\exp(z_{s,i}/T)}{\sum_j \exp(z_{s,j}/T)}
\]  
where \(q_i\) and \(p_i\) are the teacher's and student's softened probabilities for class \(i\), \(z_t\) and \(z_s\) are logits from the teacher and student, and \(T\) is the temperature.  

Then, the loss function is computed as:  
\[
\mathcal{L} = \alpha T^2 \cdot \text{KL}(q \parallel p) + (1 - \alpha) \cdot \text{CE}(y, p')
\]  
where \(\text{KL}(q \parallel p)\) is the Kullback-Leibler divergence between teacher and student distributions; \(\text{CE}(y, p')\) is cross-entropy loss between student predictions \(p' = \text{softmax}(z_s)\) and true labels \(y\); \(\alpha\) is the weight balancing distillation and cross-entropy losses; \(T^2\) compensates for gradient scaling caused by temperature. Moreover, \(z_t, z_s\) is logits from teacher and student; \(T\) is the emperature (\(T > 1\) smooths probabilities, \(T = 1\) recovers standard softmax); \(\alpha\) is the distillation weight (typically \(0.5\)); \(q, p\) is the teacher/student softened probabilities; \(y\) is the ground-truth labels. 

\paragraph{Data Distillation}
Data Distillation generates synthetic data that encapsulates the teacher's knowledge for student training. 
Synthetic data \(X_{\text{syn}}, Y_{\text{syn}}\) is generated such that training the student on it mimics training on the original dataset with the teacher. Methods include gradient matching or leveraging the teacher to label synthetic samples.

Minimize the difference between student gradients on synthetic data and teacher gradients on real data:  
\[
\mathcal{L} = \sum \left\| \nabla_\theta \mathcal{L}_{\text{syn}}(\theta) - \nabla_\theta \mathcal{L}_{\text{real}}(\theta) \right\|^2
\]  
where \(\mathcal{L}_{\text{syn}}\) is the loss on synthetic data \((X_{\text{syn}}, Y_{\text{syn}})\); \(\mathcal{L}_{\text{real}}\) is loss on real data \((X, Y)\); \(\theta\) is the model parameters.  
Moreover, \(X_{\text{syn}}, Y_{\text{syn}}\) is the synthetic inputs and labels (learned via optimization); \(\theta\) is the parameters of the student model.







\subsection{Search Strategies}


Search Strategies, often termed decoding strategies, are essential for guiding text generation in LLMs. These strategies range from simple greedy approaches to more sophisticated methods like Beam Search, Best-of-N, and Monte Carlo Tree Search.

\paragraph{Greedy Strategy}
The Greedy Search is one of the simplest decoding strategies used in LLMs. At every step, it selects the token with the highest probability as the next token in the sequence. Mathematically, this can be expressed as:

\begin{equation}
w_t = \arg\max_w P(w | w_{1:t-1})
\end{equation}
where $w_t$ represents the token chosen at time step $t$, given the previously generated tokens $w_{1:t-1}$. This method is efficient but may not always yield the most optimal or diverse output since it follows only a single path through the probability tree, potentially missing higher probability branches.

\paragraph{Beam Search}
Beam Search is an extension of the Greedy Search that maintains multiple hypotheses (or ``beams") simultaneously, allowing for a broader exploration of possible sequences. It keeps track of the top-$k$ most probable sequences at each step, where $k$ refers to the beam width. The process can be described by:
\begin{equation}
\text{Beam}(t) = \arg\max_{\text{Beam}(t-1)} \sum_{i=1}^{k} P(w_i | w_{1:t-1})
\end{equation}
Here, $\text{Beam}(t)$ denotes the set of $k$ best sequences at time step $t$. Although Beam Search often produces better results than Greedy Search, it comes at the cost of increased computational complexity.

\paragraph{Best of N}
The Best of N strategy involves generating $N$ different sequences using either Greedy or Beam Search and then selecting the best sequence based on some criterion, such as likelihood or a specific scoring function. The equation representing this approach is:
\begin{equation}
S^* = \arg\max_{S \in \{S_1, S_2, ..., S_N\}} \text{Score}(S)
\end{equation}
where $S^*$ is the selected sequence, and $\text{Score}(S)$ could be any evaluation metric applied to each of the $N$ generated sequences. This method improves diversity and quality but requires additional computational resources to generate and evaluate multiple candidates.

\paragraph{Monte Carlo Tree Search}
Monte Carlo Tree Search (MCTS) is a heuristic search algorithm that combines random sampling with tree search to explore possible actions in decision-making processes. In the context of LLMs, MCTS can be adapted to perform thought expansion and value estimation during text generation. The four main stages in each iteration of MCTS include node selection, thought expansion, greedy Monte Carlo rollout, and value update. The general procedure can be summarized as:
\begin{equation}
V(s) = \frac{1}{N} \sum_{i=1}^{N} R(s, a_i)
\end{equation}
where $V(s)$ is the estimated value of state $s$, $N$ is the number of simulations, $a_i$ are the actions sampled from policy $\pi$, and $R(s, a_i)$ is the reward obtained from taking action $a_i$ in state $s$ [[2]]. While computationally intensive, MCTS provides a robust framework for exploring complex state spaces and making informed decisions.



\subsection{Self-training}

Self-training is a semi-supervised learning technique where a model trains itself by leveraging its own predictions on unlabeled data after being initially trained on labeled data. Below, we discuss offline self-training, online self-training, and reinforced self-training with their respective equations and definitions.

\paragraph{Offline Self-training}
Offline self-training refers to the process where a model generates pseudo-labels for unlabeled data based on its predictions and then retrains itself using this augmented dataset. The training process does not involve real-time updates or feedback loops. Most existing methods iteratively update pseudo-labels and retrain the models in an offline manner \cite{zhao2023towards}.

The objective function for offline self-training can be expressed as:
\begin{equation}
    \mathcal{L}_{\text{offline}} = \mathcal{L}_{\text{labeled}} + \lambda \mathcal{L}_{\text{pseudo}},
\end{equation}
where $\mathcal{L}_{\text{labeled}}$ is the loss computed on the labeled dataset, $\mathcal{L}_{\text{pseudo}}$ is the loss computed on the pseudo-labeled dataset, and $\lambda$ is a weighting factor that balances the contributions of the two terms.

\paragraph{Online Self-training}
In contrast to offline self-training, online self-training incorporates real-time updates during the training process. This means that the model continuously refines its predictions and adapts to new observations as they arrive. Online self-training benefits from an immediate feedback loop, which allows instructors (or algorithms) to quickly address learning gaps.

The updated objective function for online self-training can be written as:
\begin{equation}
    \mathcal{L}_{\text{online}}(t) = \mathcal{L}_{\text{labeled}}(t) + \eta \mathcal{L}_{\text{unlabeled}}(t),
\end{equation}
where $t$ represents the time step, $\mathcal{L}_{\text{labeled}}(t)$ is the loss on labeled data at time $t$, $\mathcal{L}_{\text{unlabeled}}(t)$ is the loss on unlabeled data at time $t$, and $\eta$ is a dynamic weighting factor that adjusts based on the confidence of the model's predictions.

\paragraph{Reinforced Self-training}
Reinforced self-training combines reinforcement learning with self-training to improve the robustness and accuracy of the model. In this approach, the model uses reinforcement signals to refine its decision-making process while generating pseudo-labels for unlabeled data. Offline self-training includes sampling data from the initial policy for fine-tuning, which serves as a simple yet effective baseline.

The reinforcement objective in reinforced self-training can be defined as:
\begin{equation}
    \mathcal{L}_{\text{reinforced}} = \mathbb{E}_{\tau \sim \pi_\theta} \left[ \sum_{t=0}^{T} \gamma^t r_t \right] + \alpha \mathcal{L}_{\text{self}},
\end{equation}
where $\tau$ is the trajectory of states and actions sampled from the policy $\pi_\theta$, $r_t$ is the reward at time $t$, $\gamma$ is the discount factor, $\mathcal{L}_{\text{self}}$ is the self-training loss, and $\alpha$ controls the trade-off between reinforcement learning and self-training objectives \cite{trung2024reft}.









%% file: tabs/evaluate.tex
\begin{table*}[t!]
\centering
\scriptsize
\caption{Comparison of typical reasoning LLMs. RM means reward models, including rule-based (Rule), outcome reward models (ORM), and process reward models (PRM). TS and BS represent Tree Search and Beam Search.}
\label{table: evaluate}
\begin{tabular}{>{\raggedright\arraybackslash}m{0.15\textwidth} >{\raggedright\arraybackslash}m{0.14\textwidth} l >{\raggedright\arraybackslash}m{0.05\textwidth} l >
{\raggedright\arraybackslash}m{0.10\textwidth} l >{\raggedright\arraybackslash}m{0.15\textwidth} }
\hline

Models & Base LLMs & RM & Search & Learning & RL & Self-training  \\ 
\hline
DeepSeek-R1 \cite{guo2025deepseek}     &       DeepSeek-V3          &    Rule/ORM    & - &     RL /SFT/CoT  & GPRO & Offline/Reinforced \\

QwQ-32B \cite{qwq-32b-preview}     &   -  &    Rule/ORM     &  -   &   RL/SFT/CoT & - & Offline \\

QwQ-Max\footnotemark[6]    &  Qwen2.5-Max  &   Rule/ORM     & -  &     RL/SFT/CoT   & DPO & Offline \\

Claude3.7\footnotemark[7]  &  -   & - & - & RL/SFT & RLHF & - \\

LMM-R1 \cite{peng2025LMM-R1}    & Qwen2.5 & Rule/ORM & - & RL/CoT & PPO & Reinforced \\

Online-DPO-R1 \cite{zhang2025dpor1}& Qwen2.5 & Rule/ORM & - & RL/SFT/CoT & DPO & Online/Reinforced \\

Grok-3 \footnotemark[8]
&  -  &    -     &    -        &    RL/COT      &     Reinforced   & -   \\

Search-o1 \cite{li2025search}     &    QwQ-32B-Preview   &  Rule/ORM   &  RAG  &   RL/SFT/CoT   & - & - \\

Marco-o1 \cite{zhao2024marco}   &  Qwen2  & - & MCTS &   SFT/CoT & - & -  \\

gpt-o1\footnotemark[9] & - &  Rule/PRM/ORM  & MCTS  &  RL/SFT/CoT  & PPO &  Reinforced/Offline\\

Skywork-o1\footnotemark[10]   
&   Llama3.1-8B  &   PRM  &  - &  RL/SFT & Q* & -  \\
Deepseek-R1-lite\footnotemark[11]     &   DeepSeek-V2.5    &  Rule/ORM  &  - &   RL/SFT/COT &   GPRO & Offline/Reinforced   \\

LLaMA-Berry \cite{zhang2024llama}  &    LLaMA3.1 &  PRM & MCTS &  RL/COT &   DPO & Online/Reinforced \\

DRT-o1 \cite{wang2024DRT-o1}  &   LlaMA/Qwen     &  -  & - &  SFT &  - & Offline  \\

HuatuoGPT-o1 \cite{chen2024huatuogpt} & LLaMA3.1  &  Rule & Verifier & RL/SFT/CoT &PPO & Offline/Reinforced  \\

Sky-T1\footnotemark[12]
&     Qwen2.5    & - & -   &   RL/SFT & PRIME/RLOO & - \\

rStar-Math \cite{guan2025rstar}   & Qwen2.5/Phi3  & PRM &  MCTS  &   RL/SFT/CoT & - & Offline/Reinforced\\

LLaMAV-o1 \cite{Omkar2025LlamaV-o1}&  Llama-3.2 &  - &  BS  &  SFT/CoT &  - & -\\ 

S1 \cite{muennighoff2025s1}&   Qwen2.5-32B     &  - &  - &  SFT&   - & - \\

Kimi k1.5 \cite{team2025kimi} & - & Rule & - &  RL/SFT/CoT   &  RLHF  & - \\

LLaVA-CoT \cite{xu2024llava}  &  Llama3.2  & -  & BS   &   SFT/CoT     & - & -\\

STILL-3 \cite{Slow_Thinking_with_LLMs_3}  & Qwen2.5-32B  &  Rule/ORM  & - & RL/COT & RLHF & -  \\

R1-Search \cite{song2025r1} & Qwen-2.5-7B & Rule & - & RL & Reinforce++ & - \\

PRIME \cite{cui2025prime} & Qwen2.5-Math  &   ORM/PRM      & TS  & RL/SFT/CoT &  RLOO/PPO/ PRIME &  Online/Reinforced  \\

HiAR-ICL \cite{wu2024HiAR-ICL} &  
Qwen2.5 
&  PRM/ORM &  MCTS  &  CoT / ICL & - & -  \\

O1-CODER \cite{zhang2024o1-Coder}   &  Qwen2.5  &  PRM & MCTS    & SFT/CoT & - & Offline/Reinforced \\

SRA-MCTS \cite{xu2024SRA-MCTS}&  
Qwen2.5  
& PRM   &  MCTS &  RL/CoT  & - & -\\

Tinyzero \cite{tinyzero}   &  Qwen2.5 & Rule/ORM & -  &    RL /SFT/CoT & RLHF & -\\

InternLM-3 \cite{cai2024internlm2}  & -  & - & - &  RL/SFT  & Online RLHF/PPO & -  \\

VLM-R1 \cite{shen2025vlmr1} &Qwen2.5-VL & Rule/ORM& -  & RL/SFT & GPRO & - \\

SimpleRL \cite{zeng2025simplerlzooinvestigatingtamingzero}   &  -  &       Rule/ORM  & - & RL   & GRPO & - \\

Open-R1 \cite{openr1} & Qwen2.5-1.5B   &  Rule/ORM &  - &   RL/SFT &  GPRO & - \\

Demystify \cite{tong2025Demystify-longCoT} & Llama3.1/Qwen2.5-Math  & Rule/ORM &   -  &  RL/SFT /CoT  &  PPO & Reinforced \\

Gemini2.5 Pro\footnotemark[13] & Gemini  & -  & -  &  RL/CoT/SFT  &  RLHF/PPO & - \\

GLM-Z1-Air\footnotemark[14]& GLM-4-Air0414  &  - &  - &   - &  - & -  \\

Wenxin-x1\footnotemark[15]& -  &  - &  - &  -  &   RLHF/PPO  & - \\

openai-o3\footnotemark[16] & -  &  - &  - &  RL/CoT/SFT  &  -  & - \\

\hline
\end{tabular}


\end{table*}

%% file: secs/04InferenceTimeScalingLaw.tex
\input{imgs/Roadmap_to_reasoning_LLM}

\begin{figure}[t]
\centering
\includegraphics[width=0.95\columnwidth]{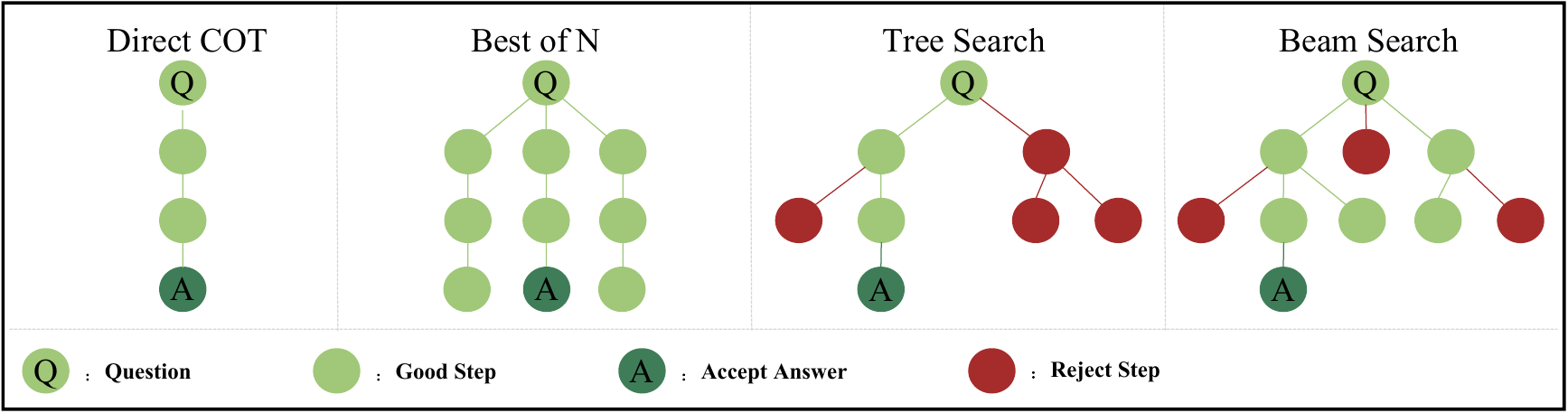}
\caption{The search algorithms for test-time scaling}
\label{fig:timeline}
\end{figure}

\section{Test-Time Scaling}
\label{sect:test-time scaling}
In this section, we present the studies about test-time scaling, which dynamically adjusts computation based on task complexity (Table \ref{table: Inference time scaling}). We split them into search and sampling, and dynamic verification mechanism.
\input{tabs/Summary_of_InferenceTime}

\subsection{Search and Sampling}
\label{sect: search and sampling}

\subsubsection{Search}
Search methods enhance model performance by systematically exploring reasoning paths. Two primary approaches exist: Beam Search and Monte Carlo Tree Search (MCTS). 
This section examines recent advancements in both methodologies.

\paragraph{Beam Search}
Beam search provides an effective balance between computational efficiency and generation quality by preserving a fixed number of the highest-scoring candidate paths (the beam width) at each step. Unlike exhaustive search methods, it selects candidate paths based on probabilistic scoring and employs pruning strategies to promote generation quality while significantly reducing computational overhead.

The LLaVA-O1 framework \cite{xu2024llava} introduces a novel stage-level beam search approach, structuring multimodal reasoning into four distinct phases (summary, caption, reasoning, conclusion) delineated by dedicated tags. In contrast to traditional token-level or sentence-level beam search, this framework incorporates three core innovations: (1) the generation of multiple candidate paths at each stage; (2) autonomous selection of high-scoring paths within each stage; and (3) the propagation of high-quality intermediate results across stages. This structured approach enables smaller models, such as LLaVA-O1, to outperform larger proprietary counterparts (e.g., GPT-4o-mini) on systematic reasoning tasks, thereby demonstrating the effectiveness of beam search when its application is tailored to domain-specific cognitive workflows.

\paragraph{Monte Carlo Tree Search}

Silver et al. \cite{silver2017mastering} significantly advanced Monte Carlo Tree Search (MCTS) with the introduction of AlphaZero, a framework integrating MCTS principles with deep reinforcement learning via self-play. By replacing traditional domain-specific heuristics with neural networks, AlphaZero achieved superhuman performance in complex games like chess and Go, representing a landmark development in MCTS methodology. Building upon this foundation, Zhao et al. \cite{zhao2411marco} proposed Marco-O1, a method that enhances action selection by decomposing reasoning steps into smaller sequences of 32 or 64 tokens. 
This strategy facilitates a more granular exploration of the search space compared to conventional step-by-step action techniques. Jiang et al. \cite{jiang2024technical} introduced a triple-optimization strategy specifically for mathematical reasoning, comprising dynamic thresholding for global leaf selection, self-consistency-enhanced simulation to filter inconsistencies, and pre-expansion mechanisms for more comprehensive tree construction. Further refining MCTS, the REBASE framework \cite{wu2024inference} improves node evaluation using Policy-guided Rollout Model strategies; it employs softmax-normalized reward scoring and reward-weighted sampling instead of traditional rollouts, enabling efficient search tree navigation even with smaller models. Moreover, rStar-Math \cite{guan2025rstar} presented a code-verified chain-of-thought synthesis method featuring automated Python validation. This approach incorporates process preference modeling using MCTS-derived Q-values and utilizes a four-round self-evolution framework that co-evolves policy Smaller Language Models through curriculum-guided exploration. Complementing these efforts, Tree of Thoughts (ToT) \cite{yao2023treethoughtsdeliberateproblem} extends MCTS principles to enable multi-path reasoning via backtracking mechanisms and the construction of complex exploration spaces, endowing language models with capabilities for human-like, stepwise problem-solving. Collectively, these innovations substantially broaden the applicability of MCTS to complex reasoning domains by improving search granularity, integrating robust verification mechanisms, and achieving an adaptive balance between exploration and exploitation.

\subsubsection{Sampling}

Sampling in generative AI entails producing multiple output candidates from identical initial conditions (e.g., prompts) and then strategically aggregating them using verification mechanisms. Two primary paradigms emerge: Majority Vote, which selects the most frequent valid answers, and Best-of-N, which leverages reward models to identify optimal candidates. Brown et al. \cite{brown2024large} laid the foundation with scaling laws, demonstrating that repeated sampling exponentially expands problem-solving coverage (with $\log c \approx a \cdot k^{b}$) while highlighting cost-efficiency tradeoffs. Their results indicate that extensive sampling from weaker models can be more economical than a single attempt using stronger models. Xie et al. \cite{xie2025sana} extended these findings to multimodal settings, revealing that sampling multiplicity outperforms simply increasing denoising steps in vision-language models for artifact correction. Addressing efficiency, Sun et al. \cite{sun2024fast} introduced the Speculative Rejection algorithm, which achieves 16$\times$--32$\times$ computational gains by dynamically terminating unpromising paths, all while maintaining performance comparable to the Best-of-N approach. Muennighoff et al. \cite{muennighoff2025s1} explored computational budget control via token limits, enforcing early answer generation through forced termination tokens or enabling extended reasoning with suppression mechanisms. However, their rejection sampling implementations showed only limited success. In a different vein, Feynman-Kac (FK) Steering \cite{singhal2025general} presents a particle-based sampling method for diffusion models that adaptively resamples trajectories using intermediate rewards. By prioritizing high-reward paths during generation, this method achieves controllable outputs across modalities and outperforms standard sampling approaches in rare-event tasks. A critical challenge remains in non-automatable domains that require hybrid verification strategies. For instance, coding applications often combine reward models with test suites \cite{brown2024large}, while mathematical reasoning relies heavily on learned precision scorers \cite{xie2025sana}.

\subsubsection{Long to Short CoT}
Efforts to optimize long COT reasoning in LLMs have led to diverse innovative approaches aimed at improving efficiency while maintaining reasoning performance. The OverThink framework \cite{kumar2025overthinkslowdownattacksreasoning} highlights vulnerabilities in reasoning LLMs, demonstrating how slowdown attacks can disrupt inference efficiency by injecting decoy reasoning problems. To counter such inefficiencies, methods like LightThinker propose dynamically compressing intermediate reasoning steps, achieving faster inference on complex tasks with minimal performance trade-offs \cite{zhang2025lightthinker}. Similarly, the TokenSkip strategy \cite{xia2025tokenskip} enables selective skipping of less critical tokens, providing controllable CoT compression, while a token-budget-aware framework adjusts the number of reasoning tokens based on task complexity, optimizing resource allocation \cite{han2024token}. Other frameworks, such as Chain of Draft \cite{xu2025chain}, focus on generating concise yet informative intermediate outputs to accelerate reasoning, and the InftyThink paradigm \cite{yan2025inftythink} restructures reasoning into iterative processes with intermediate summaries, significantly reducing computational demands. Additionally, the Sketch-of-Thought (SoT) framework \cite{aytes2025sketch} integrates cognitive-inspired strategies with linguistic constraints to minimize token usage while preserving accuracy. Collectively, these approaches address the challenges of long CoT reasoning, enabling more efficient and scalable reasoning systems for practical applications.

\subsection{Dynamic Verification Mechanism}
\label{sect:  dynamic verification mechanism}
\subsubsection{Verification Guided Strategy}

Verification Guided Strategy is a test-time optimization method that generates multiple candidates and selects the best output using domain-specific verifiers to enhance quality without modifying base model parameters.
CoRe \cite{zhu2022solving} introduces a dual-system cognitive framework that decouples reasoning into generation and verification phases, employing MCTS for inference-time feedback. 
Zhao et al. \cite{zhao2025samplescrutinizescaleeffective} revealed that expanding sampling-based search methods with self-verification strategies significantly improve reasoning capabilities, leveraging both implicit expansion phenomena and comparative verification mechanisms.
Extending these principles to cross-modal applications, Llasa \cite{ye2025llasa} implements a Llama-based TTS system that scales train-time compute for improved speech naturalness while incorporating inference-time verification search to enhance emotional expressiveness and timbre consistency. 
Similarly, in specialized domains, Ni et al. \cite{ni2023leverlearningverifylanguagetocode} trained validators to determine the correctness of LLM sampling, and combined the validation score with the probability of LLM generation to re-rank the sampling results, enhancing the ability of LLM to generate code.

\subsubsection{Self-Refine Strategy}

Based on evaluation outcomes, the model identifies errors or deficiencies and initiates corrective actions, such as self-refinement or regeneration, to enhance the quality of its output \cite{weng2023largelanguagemodelsbetter,yao2023treethoughtsdeliberateproblem,xie2023selfevaluationguidedbeamsearch,gero2023selfverificationimprovesfewshotclinical,shinn2023reflexionlanguageagentsverbal}.

Intrinsic Evaluation and Confidence Estimation:
Models can be trained or prompted to intrinsically assess the reliability of their generated content. 
For instance, Yao et al. \cite{yao2024learning} introduced a method for measuring step-wise confidence based on generation logits, which considers both intra-step and inter-step confidence levels to identify potential weaknesses within the reasoning chain. SelfCheck \cite{miao2023selfcheck} employs a multi-stage process wherein the model verifies the conditional correctness of its reasoning steps and subsequently generates an overall confidence score. 

Step-wise Verification and Error Localization:
Errors within reasoning processes often accumulate incrementally, making verification and correction at intermediate stages crucial. 
SelfCheck \cite{miao2023selfcheck} is specifically designed to identify the precise steps where errors occur within a reasoning chain. 
Yao et al. \cite{yao2024learning} identified the earliest erroneous step by calculating step-wise confidence scores and subsequently restarting the reasoning process from the last validated correct step. 
Xie et al. \cite{xie2023self} proposed Guided Beam Search that conducts self-assessment at each step of the beam search algorithm to guide the selection of promising reasoning paths.

%% file: imgs/Roadmap_to_reasoning_LLM.tex
\tikzstyle{my-box}=[
    rectangle,
    draw=hidden-black,
    rounded corners,
    text opacity=1,
    minimum height=1.5em,
    minimum width=5em,
    inner sep=2pt,
    align=center,
    fill opacity=.5,
]
\tikzstyle{leaf}=[
    my-box, 
    minimum height=1.5em,
    fill=hidden-blue!90, 
    text=black,
    align=left,
    font=\normalsize,
    inner xsep=2pt,
    inner ysep=4pt,
]
\begin{figure*}[t]
    \vspace{-2mm}
    \centering
    \resizebox{\textwidth}{!}{
        \begin{forest}
            forked edges,
            for tree={
                grow=east,
                reversed=true,
                anchor=base west,
                parent anchor=east,
                child anchor=west,
                base=left,
                font=\large,
                rectangle,
                draw=hidden-black,
                rounded corners,
                align=left,
                minimum width=4em,
                edge+={darkgray, line width=1pt},
                s sep=6pt,
                inner xsep=2pt,
                inner ysep=3pt,
                line width=0.8pt,
                ver/.style={rotate=90, child anchor=north, parent anchor=south, anchor=center},
            },
            where level=1{text width=7.6em,font=\normalsize,}{},
            where level=2{text width=9.4em,font=\normalsize,}{},
            where level=3{text width=12em,font=\normalsize,}{},
            where level=4{text width=30em,font=\normalsize,}{},
            where level=5{text width=30em,font=\normalsize,}{},
            [
                Roadmap to Reasoning LLM, ver
                [
                    Test-Time \\ Scaling ~(\S\ref{sect:test-time scaling})
                    [
                        Search and \\ Sampling ~(\S\ref{sect: search and sampling})
                        [
                            Beam Search
                            [
                                {E.g.,} 
                                LLaVA-o1 \cite{xu2024llava}
                                , leaf, text width=44.6em
                            ]
                        ]
                        [
                            Monte Carlo Tree \\Search
                            [
                                {E.g.,} 
                                Marco-o1 \cite{zhao2411marco}{,}
                                STILL-1  \cite{jiang2024technical}{,}
                                REBASE \cite{wu2024inference}{,}
                                ToT \cite{yao2023treethoughtsdeliberateproblem} 
                                , leaf, text width=44.6em
                            ]
                        ]
                        [
                            Sampling
                            [
                                {E.g.,} 
                                Brown et al. \cite{brown2024large}{,}
                                SANA1.5 \cite{xie2025sana}{,}
                                Speculative Rejection \cite{sun2024fast}{,}
                                S1 \cite{muennighoff2025s1}{,}
                                Feynman-Kac Steering \cite{singhal2025general}
                                , leaf, text width=44.6em
                            ]      
                        ]
                        [
                            Long to Short CoT
                            [
                                {E.g.,} 
                                {
                                OverThink \cite{kumar2025overthinkslowdownattacksreasoning}{,}
                                LightThinker \cite{zhang2025lightthinker}{,}
                                TokenSkip \cite{xia2025tokenskip}{,}
                                TALE \cite{han2024token}{,}}\\
                                COD \cite{xu2025chain}{,}
                                InftyThink \cite{yan2025inftythink}{,}
                                SoT  \cite{aytes2025sketch}
                                , leaf, text width=44.6em
                            ]
                        ]
                    ]
                    [
                        Dynamic Verification\\ Mechanism 
                        (\S\ref{sect:  dynamic verification mechanism})
                        [
                            Verification Guided \\Strategy
                            [
                                {E.g.,} 
                                CoRe \cite{zhu2022solving}{,}
                                SCoRe \cite{kumar2024training}{,}
                                Zhao et al. \cite{zhao2025samplescrutinizescaleeffective}{,}
                                Llasa \cite{ye2025llasa}{,}
                                HuatuoGPT-o1 \cite{chen2024huatuogpt}
                                , leaf, text width=44.6em
                            ]
                        ]
                        [
                            Self-Refine Strategy
                            [
                                {E.g.,} 
                                Self-Verification \cite{weng2023largelanguagemodelsbetter}{,}
                                ToT \cite{yao2023treethoughtsdeliberateproblem}{,}
                                AutoMathCritique  \cite{xi2024enhancing}{,}
                                Gero et al. \cite{gero2023selfverificationimprovesfewshotclinical}{,}
                                Reflexion \cite{shinn2023reflexionlanguageagentsverbal}
                                 , leaf, text width=44.6em 
                            ]
                        ]
                    ] 
                ]
                [
                    Reinforced\\ Learning ~(\S\ref{sect: reinforced learning})
                    [
                        Policy Network~(\S\ref{sect: policy network})
                        [
                            Training Data Acquisition
                            [
                                {E.g.,} 
                                T1 \cite{hou2025advancing}{,}
                                SCoRe \cite{kumar2024training}{,}
                                LLaVA-O1 \cite{xu2024llava}{,}
                                ReST \cite{gulcehre2023reinforced}{,}
                                SPIN \cite{chen2024self}{,}
                                DeepseekMath \cite{shao2024deepseekmath}{,}\\
                                MiniLLM \cite{gu2024minillm}{,}
                                Phi-4 \cite{abdin2024phi}
                                , leaf, text width=44.6em
                            ]
                        ]
                        [
                            Multi-Stage Training\\ Strategies
                            [
                                {E.g.,}
                                Deepseek-R1 \cite{guo2025deepseek}{,}
                                BOLT \cite{pang2025bolt}{,}
                                Math-shepherd \cite{wang2024math}{,}
                                o1-Coder \cite{zhang2024o1}{,}
                                MiniLLM \cite{gu2024minillm}{,}\\
                                Recursive introspection \cite{qu2024recursive}
                                , leaf, text width=44.6em
                            ]
                        ]    
                    ]
                    [
                        Reward Design ~(\S\ref{sect: reward design})
                        [
                            Supervision-Based Methods
                            [
                                {E.g.,} 
                                Lightman et al. \cite{lightman2023let}{,}
                                AtomThink \cite{xiang2024atomthinkslowthinkingframework}{,}
                                BOLT \cite{pang2025bolt}{,}
                                Zeng et al. \cite{zeng2024scaling}{,}
                                SPIN \cite{chen2024self}{,}\\
                                OpenRFT \cite{zhang2024openrft}{,}
                                Open-Reasoner-Zero \cite{OpenReasonerZero2025}{,}
                                Deepseek-R1 \cite{guo2025deepseek}
                                , leaf, text width=44.6em
                            ]
                        ]
                        [
                            Reward Source-Based Models
                            [
                                {E.g.,} 
                                Phi-4 \cite{abdin2024phi}{,}
                                PPO \cite{schulman2017proximal}{,}
                                HuatuoGPT-o1 \cite{chen2024huatuogpt}{,}
                                PPO-max \cite{zheng2023secrets}{,}
                                DPO \cite{rafailov2024direct}{,}
                                 ReST \cite{gulcehre2023reinforced}{,}
                                  SCoRe \cite{kumar2024training} 
                                , leaf, text width=44.6em
                            ]
                        ] 
                    ]
                    [
                        Self-Evolution ~(\S\ref{sect: self-evolution})
                        [
                            Self-Evaluation and Feedback
                            [
                                 {E.g.,} 
                                Self-Rewarding Correction \cite{xiong2025self}{,}
                                Selfcheck \cite{miao2023selfcheck}{,}
                                rstar-Math \cite{guan2025rstar}{,}
                                LECO \cite{yao2024learning}{,}\\
                                SELF-REFINE \cite{madaan2024self}{,}
                                Recursive introspection \cite{qu2024recursive}
                                , leaf, text width=44.6em
                            ]
                        ]
                        [
                            Reinforcement Learning \\ and Self-Training
                            [
                                 {E.g.,} 
                                STaR \cite{zelikman2024star}{,}
                                Kwai-STaR  \cite{lu2024kwai}{,}
                                Quiet-STaR \cite{zelikman2024quiet}{,}
                                 SCoRe \cite{kumar2024training}{,}
                                ReST \cite{gulcehre2023reinforced}{,}
                                  SPIN \cite{chen2024self} 
                                 , leaf, text width=44.6em
                             ]
                        ]
                    ]
                ]
                [
                    Slow Thinking ~(\S\ref{sect: slow thinking})
                    [
                        Long CoT (\S\ref{sect: long cot})
                        [
                            Data Distillation
                            [
                                {E.g.,} 
                                  Wu et al. \cite{wu2024comparativestudyreasoningpatterns}{,}
                                Deepseek-R1~\cite{guo2025deepseek}{,}
                                 AtomThink \cite{xiang2024atomthinkslowthinkingframework}{,}
                                 CoT-Valve \cite{ma2025cot}{,}\\
                                Distilling System 2 into System 1 \cite{yu2024distilling21}
                                , leaf, text width=44.6em
                            ]
                        ]
                        [
                            Long-Context Extension \\ \& Improvement.
                            [
                                {E.g.,} 
                                Kimi k1.5 \cite{team2025kimi}{,}
                                Marco-o1 \cite{zhao2024marco}{,}
                                multiple iterations in ICL \cite{yang2023iterative}{,}
                                Phi-4 \cite{abdin2024phi}
                                ,leaf, text width=44.6em
                            ]
                        ]
                        [
                            Implicit Reasoning.
                            [
                                {E.g.,}
                                Kimi k1.5 \cite{team2025kimi}{,}
                                Deng et al. \cite{deng2024explicit}{,}
                                MAD \cite{liang2023encouraging}
                                ,leaf, text width=44.6em
                            ]
                        ]
                        [
                            Reflection and \\Backtracking.
                            [
                                {E.g.,}
                                Deepseek-R1~\cite{guo2025deepseek}{,}
                                RedStar \cite{xu2025redstar}{,}
                               Distilling System 2 into System 1 \cite{yu2024distilling21}{,}\\
                               CKT-MHA \& CL-vMF mechanism \cite{qi2024interactive}{,}
                               Min et al. \cite{min2024imitateexploreselfimprovereproduction}{,}
                                Gan et al. \cite{gan2025rethinkingexternalslowthinkingsnowball}
                                ,leaf, text width=44.6em
                            ]
                        ]
                    ]
                    [
                        Hierarchical Reasoning \\ (\S\ref{sect: hierarchical reasoning frameworks})
                        [
                            Explicit Structures
                            [
                                {E.g.,}
                                ReasonFlux \cite{2025arXiv250206772Y}{,}
                                Li et al. \cite{li2025search}
                                ,leaf, text width=44.6em
                            ]
                        ]
                        [
                            Agentic Systems
                            [
                                {E.g.,}
                                MALT \cite{motwani2025maltimprovingreasoningmultiagent}{,}
                                OctoTools \cite{lu2025octotoolsagenticframeworkextensible}{,}
                                Agentic Reasoning \cite{wu2025agenticreasoningreasoningllms}
                                ,leaf, text width=44.6em
                            ]
                        ]
                        [
                            Dynamic Control Mechanism
                            [
                                {E.g.,}
                                MixLLM \cite{wang2025mixllmdynamicroutingmixed}{,}
                                AdaptiveStep \cite{liu2025adaptivestepautomaticallydividingreasoning}
                                ,leaf, text width=44.6em
                            ]
                        ]
                        [
                            Latent Space Manipulations
                            [
                                {E.g.,}
                               Yang et al. \cite{yang2023iterative}{,}
                                LTMs \cite{kong2025scalable}{,}
                                PEARL \cite{chen2025pearlpermutationresilientllms}{,}
                                ThoughtProbe \cite{wang2025thoughtprobeclassifierguidedthoughtspace}
                                ,leaf, text width=44.6em
                            ]
                        ]
                    ]
                    [
                        Hybrid Thinking \\(\S\ref{sect: Hybrid Thinking model})
                        [
                            Guided Search
                            [
                                {E.g.,}
                                HDFlow \cite{yao2024hdflow}{,}
                                Dualformer \cite{su2024dualformer}{,}
                                HaluSearch \cite{zhang2024fast}{,}
                                Q* \cite{deng2024q}{,}
                                Mulberry \cite{yao2024mulberryempoweringmllmo1like}{,}
                                FoO \cite{nair2025flowofoptionsdiversifiedimprovedllm}
                                ,leaf, text width=44.6em
                            ]
                        ]
                        [
                            Adaptive Control
                            [
                                {E.g.,}
                                DAST \cite{shen2025dastdifficultyadaptiveslowthinkinglarge}{,}
                                Entro-duction \cite{zhang2025entropybasedexplorationconductionmultistep}{,}
                                SIFT \cite{zeng2025siftgroundingllmreasoning}
                                ,leaf, text width=44.6em
                            ]
                        ]
                        [
                            Specialized \\ Architectures
                            [
                                {E.g.,}
                                Talker-Reasoner \cite{christakopoulou2024agents}{,}
                                FS-GEN \cite{zhang2024fast}{,}
                                SYMBOLIC-MoE \cite{chen2025symbolicmixtureofexpertsadaptiveskillbased}{,}
                                Lemmanaid \cite{alhessi2025lemmanaidneurosymboliclemmaconjecturing}
                                ,leaf, text width=44.6em
                            ]
                        ]
                        [
                            Tailored Training
                            [
                                {E.g.,}
                                {RELAY \cite{yu2025enhancingautoregressivechainofthoughtloopaligned}{,}
                                Mix Distillation \cite{li2025smallmodelsstrugglelearn}{,}
                                MoBA \cite{lu2025mobamixtureblockattention}{,}
                                B-STaR \cite{zeng2025bstarmonitoringbalancingexploration}{,}
                                FAST \cite{sun2024visual}{,}
                                Hu et al. \cite{hu2023tree}{,}}\\
                                MathFusion \cite{pei2025mathfusionenhancingmathematicproblemsolving}{,}
                                Meta-RFFT \cite{hu2025traininglargelanguagemodels}
                                ,leaf, text width=44.6em
                            ]
                        ]
                    ]
                ]
            ]
        \end{forest}
    }
    \vspace{-5mm}
    \caption{Roadmap to Reasoning LLM.}
    \label{fig:task_taxonomy}
    \vspace{-4mm}
\end{figure*}

%% file: tabs/Summary_of_InferenceTime.tex
\begin{table*}[t!]
\centering
\small
\caption{ Summary of Inference. L2S means Long to Short CoT. VG and SR mean Verification-Guided and Self-Refine strategies. }
\label{table: Inference time scaling}
\setlength{\tabcolsep}{1.0mm}{
\begin{tabular}{l >{\raggedright\arraybackslash}m{0.17\textwidth} >{\raggedright\arraybackslash}m{0.17\textwidth}|ccccc}
\hline
Methods     & Search and Sampling         & Dynamic Verification   & MATH & AIME 2024 & GPQA & MMLU & GSM8K\\ \hline
Deepseek R1 \cite{guo2025deepseek} & Search & VG&-& 79.8 &	71.5 &	90.8& 	- \\ 
Marco-o1 \cite{zhao2411marco}& Search & VG  & - &-  & - & - &-\\ 
LLaVA-o1 \cite{xu2024llava} & Search &  VG & - &-  &  -& - &-\\ 
STILL-2 \cite{min2024imitate} &  Search&  VG   & - &  46.9& 56.1 & - &-\\ 
SANA1.5 \cite{xie2025sana} & Sampling  & SR  &- &-  &  -&  -&-\\ 
S1 \cite{muennighoff2025s1}  & Sampling & SR  &-  &56.7  &59.6 & - &-\\ 
CoRe \cite{zhu2022solving} &Search  &VG   & - &-  & -& - &-\\ 
SCoRe \cite{kumar2024training}  &  Search&  VG &  64.4&-  & - &  -&-\\ 
STaR \cite{zelikman2022star}  &  -&  SR &  -&-  & - &  -&10.7\\ 
V-STaR \cite{hosseini2024v} &  Sampling &SR &-  &-  &- & - &63.2\\ 
Quiet-STaR \cite{zelikman2024quiet} &  - &  SR &-  &-  &-&  -&48.0\\ 
HuatuoGPT-o1 \cite{chen2024huatuogpt}& L2S &-  &- & - & 66.5& 82.8 &-\\ 
OverThink \cite{kumar2025overthinkslowdownattacksreasoning}& L2S &-  &-  & - & -&  -&-\\ 
Inftythink \cite{yan2025inftythink}& L2S &SR   &62.5  &65.6&- & - &-\\ 
LightThinker \cite{zhang2025lightthinker}& L2S & SR  & - &-  &36.3 &60.5  &90.1\\ 
\hline
\end{tabular}
}
\end{table*}

%% file: secs/05ReinforcedLearningReasoning.tex
\begin{figure}[t]
\centering
\includegraphics[width=0.95\columnwidth]{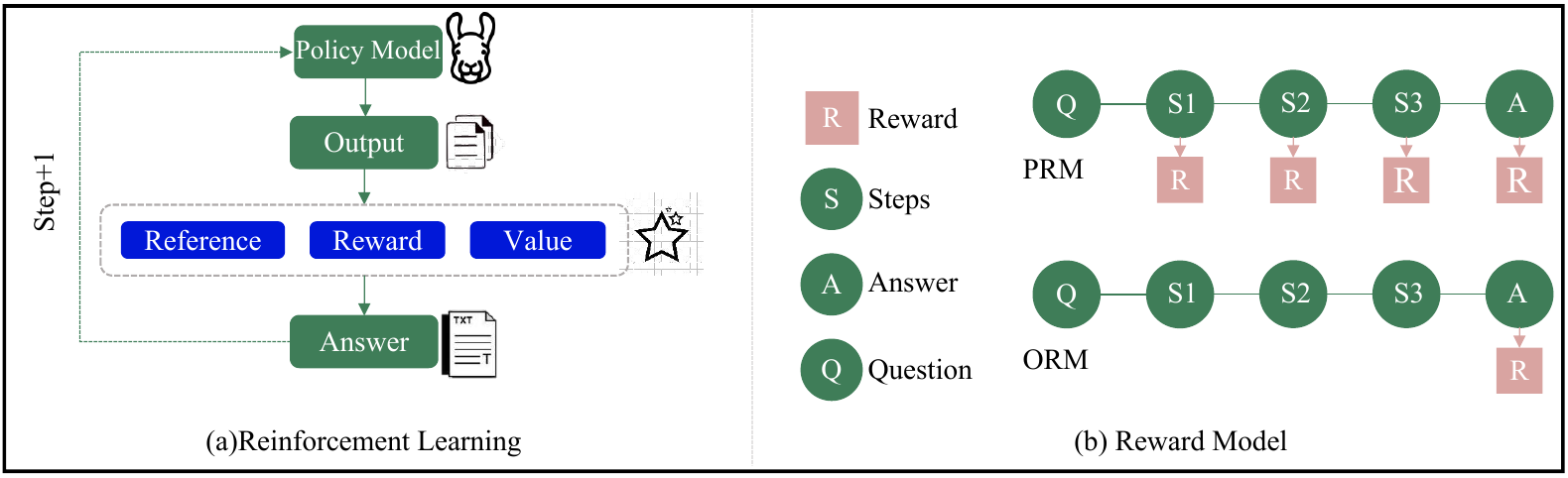}
\caption{The reinforcement learning framework and reward model}
\label{fig:timeline}
\end{figure}

\section{Reinforced Learning}
\label{sect: reinforced learning}
In this section, we summarize the related studies of reinforced learning for reasoning LLMs, including policy network, reward design and self-evolution (Table \ref{table: REINFORCED LEARNING}).

\input{tabs/summary_of_REINFORCED_LEARNING}

\subsection{Policy Network}
\label{sect: policy network}


The policy of reasoning with large models \cite{zeng2024scaling} is crucial for enhancing their reasoning capabilities. Current research mainly focuses on training data acquisition and multi-stage training strategies. 

\subsubsection{Training Data Acquisition}

Training data acquisition Strategies are essential approaches designed to address the critical challenge of limited data availability during the initial training phase of reasoning models. These strategies primarily consist of two complementary approaches: Data Synthesis and Augmentation, and Transfer Learning. 

\paragraph{Data Synthesis and Augmentation}
To address data scarcity during the cold start phase, various data synthesis and augmentation methods have been developed. These approaches generate synthetic data to supplement real data, increase diversity, and improve model generalization. For instance, Hou et al. \cite{hou2025advancing} initialized LLMs with synthetic chain-of-thought data, integrating trial-and-error and self-verification.
Kumar et al. \cite{kumar2024training} prompted base models to generate self-correction trajectories.
Xu et al. \cite{xu2024llava} created datasets with detailed reasoning processes.
Methods in Gulcehre et al. \cite{gulcehre2023reinforced} and Chen et al. \cite{chen2024self} expanded training sets by generating data from current policies.

\paragraph{Transfer Learning}
Transfer learning is widely applied in strategy initialization for reasoning models. By leveraging existing model foundations or knowledge from related domains, it reduces dependence on new data and accelerates training for new tasks \cite{theodoris2023transfer}. Examples include:
 Shao et al. \cite{shao2024deepseekmath} initialized mathematical reasoning models based on code-trained models.
Gu et al. \cite{gu2024minillm} used pre-trained models as foundations for transfer learning.
Abdin et al. \cite{abdin2024phi} transferred knowledge and capabilities from previous models.

\subsubsection{Multi-Stage Training Strategies}

Multi-Stage Training Strategies represent a systematic approach to developing reasoning capabilities in large models through sequential refinement stages. This approach consists of two primary phases: the Cold Start Fine-Tuning Stage and the Rejection Sampling and Supervised Fine-Tuning Stage. Together, these multi-stage strategies create a progressive learning pathway that systematically builds and refines reasoning abilities, ultimately resulting in more robust and capable reasoning models.

\paragraph{Cold Start Fine-Tuning Stage}
The cold start fine-tuning stage typically uses small amounts of high-quality reasoning data to preliminarily fine-tune base models. This helps models quickly develop effective reasoning frameworks, generating reasonable outputs early in training and laying foundations for subsequent reinforcement learning or further training. Examples include:
Guo et al. \cite{guo2025deepseek} fine-tuned models with collected cold start data.
The LongCoT Bootstrapping stage in Pang et al.\cite{pang2025bolt} used a few in-context examples to guide LongCoT data generation.
Methods in Wang et al. \cite{wang2024math} and Zhang et al. \cite{zhang2024o1} also used specific data for initial fine-tuning.

\paragraph{Rejection Sampling and Supervised Fine-Tuning Stage}
This stage collects high-quality reasoning data through rejection sampling and other methods, filters out low-quality reasoning chains, and uses the refined data for further supervised fine-tuning. This enhances models' comprehensive capabilities by incorporating knowledge from other domains. Examples include:
\cite{guo2025deepseek} collects SFT data with rejection sampling after reinforcement learning convergence.
\cite{gu2024minillm} uses rejection sampling to gather high-quality reasoning data for supervised fine-tuning.
\cite{qu2024recursive} selects high-quality reasoning data with rejection sampling for supervised fine-tuning.


\subsection{Reward Design}
\label{sect: reward design}

In LLMs engineered for sophisticated reasoning, reinforcement learning plays a pivotal role, and reward models (RMs) are fundamental to the success of RL frameworks\cite{zeng2024scaling}.

The primary function of the reward model is to provide evaluative feedback on either the model's reasoning process or its final output, thereby steering the policy model's optimization towards desired behaviors, such as generating correct, coherent, and reliable inference steps.
Numerous studies—including \cite{ElKishky2025CompetitivePW,guo2025deepseek,team2025kimi,hou2025advancing,chen2024huatuogpt} alongside open-source projects like \cite{wang2024openrt} have underscored the efficacy of utilizing RL and meticulously designed reward mechanisms to foster and improve LLM reasoning abilities. For instance, El-Kishky et al. \cite{ElKishky2025CompetitivePW} demonstrated how scaled RL applications that are predicated on effective reward signals enabled models like o3 to achieve superhuman performance in programming competitions, while \cite{guo2025deepseek} showed that RL alone, guided by reward models, can effectively cultivate the reasoning potential within models.


\subsubsection{Supervision-based Methods} The supervision-based methods consists of process supervision, outcome supervision and hybrid model.


\paragraph{ Process Supervision}

Process Reward Models (PRMs), also known as Process Supervision, operate on the principle of evaluating and assigning scores to each step or intermediate state within the reasoning process, rather than solely focusing on the final answer. This fine-grained approach to supervision has proven particularly effective for complex, multi-step reasoning tasks, such as mathematical problem-solving. By rewarding correct intermediate reasoning steps, rather than only the final outcome, PRMs guide the model towards learning more reliable and interpretable reasoning pathways. This reduces the likelihood of models reaching correct answers via flawed reasoning, thereby enhancing the alignment of the model's reasoning behavior \cite{lightman2023let,wang2024openrt}. 

Research by Lightman et al. \cite{lightman2023let} explicitly demonstrated that process supervision significantly outperforms outcome supervision on the MATH dataset, highlighting its benefits in terms of reliability, alignment advantages. 
Several advanced reasoning models, such as OpenR \cite{wang2024openrt}, which is inspired by OpenAI's o1 series, Open-O1 \cite{openr1}, and AtomThink \cite{xiang2024atomthinkslowthinkingframework}, explicitly leverage PRMs to improve their step-by-step reasoning and self-verification capabilities. 
Moreover, researchers are exploring methods for constructing PRMs without requiring manual annotation, such as Math-shepherd\cite{wang2024math}, which trains models using automatically generated process supervision data, and leveraging critique models to provide feedback on reasoning steps \cite{xi2024enhancing}.

\paragraph{ Outcome Supervision}

In contrast to Process Reward Models, Outcome Reward Models (ORMs), also known as outcome supervision, provide reward signals based solely on the correctness or quality of the final task output. Examples include assessing whether the final answer to a mathematical problem is correct \cite{lightman2023let} or whether generated code passes predefined test cases \cite{ElKishky2025CompetitivePW}. The advantage of this approach lies in its relative implementation simplicity, particularly for tasks yielding clear, binary (correct/incorrect), or otherwise quantifiable outcomes.

However, a primary limitation of outcome supervision is the sparsity of its signal; as feedback is provided only upon completion, errors within intermediate reasoning steps may remain undetected. More critically, ORMs can incentivize models to learn ``shortcut solutions," whereby correct answers are obtained serendipitously through non-rigorous or even flawed reasoning paths. This tendency undermines the model's generalization capabilities and its reliability on unseen problems. Despite these limitations, ORMs are still employed in practice, sometimes serving as baseline models for comparative analysis \cite{wang2024math} or integrated with other components within specific frameworks, such as evaluating leaf nodes in tree search algorithms \cite{jiang2024technical} or combined with techniques like Direct Preference Optimization (DPO) during training \cite{pang2025bolt}. Furthermore, studies investigating test-time computational scaling laws often rely on outcome-level evaluation metrics \cite{liu2025can}.

\paragraph{ Hybrid Model}
To integrate the fine-grained guidance benefits of process supervision with the goal alignment capabilities of outcome supervision, researchers have begun exploring strategies that combine these two supervisory signals. These hybrid approaches aim to leverage Process Reward Models for detailed procedural guidance while employing Outcome Reward Models to ensure the correctness of the final outcome.

For instance, within certain search-based reasoning frameworks, such as reward-guided tree search, process evaluation may guide the search direction, while outcome verification is employed to ultimately evaluate and select complete reasoning paths \cite{xi2024enhancingllmreasoningcritique}. This approach is theorized to offer a more robust supervisory signal, balancing reasoning rigor with goal attainment.  
Anthony et al. \cite{599c7963601a182cd2638264} explicitly discussed the concept of integrating process and outcome reward models. 
In the context of medical reasoning tasks, HuatuoGPT-o1 \cite{chen2024huatuogpt} first employs an outcome-oriented validator to guide the search for complex reasoning trajectories and subsequently applies reinforcement learning based on rewards from this validator. Numerous approaches employing search algorithms like Monte Carlo Tree Search, such as Llama-berry 
 \cite{zhang2024llama}, which incorporates Path-level Process Reward Models (PPRM) for global path evaluation and rStar-Math \cite{guan2025rstar}, combining a Process Preference Model (PPM) with code verification, similarly exemplify the integration of process-level exploration with final outcome or preference assessment. The STILL-1 framework  \cite{jiang2024technical}, which utilizes an ORM to guide tree search, further illustrates this combination of process and outcome supervision.


\subsubsection{ Reward Source-based Models}
We also split the existing reward models into rule-based reward models and preference learning based on reward sources.

\paragraph{ Rule-Based Reward Models}
Rule-based reward models rely on pre-defined rules, heuristics, or automated validators, which can be programmatically executed, to generate reward signals \cite{abdin2024phi}. 
These rules are often based on various criteria, including logical consistency, mathematical equivalence, the correctness of code execution outcomes, and adherence to specific constraints \cite{guan2025rstar,wang2024math}. 
Historically, such rule-based reward systems were commonly employed in game environments like chess and Go. For instance, AlphaGo utilized pre-defined rules to assess the quality of game moves and refine its strategy. Owing to their inherent advantages, such as high objectivity, interpretability, scalability, and the capacity to significantly reduce reliance on large-scale, costly human annotations, rule-based reward models have also found widespread application in large model training. For example, Open-Reasoner-Zero \cite{OpenReasonerZero2025} employed a simple binary rule-based reward determined by the correctness of the final answer. Similarly, HuatuoGPT-o1 \cite{chen2024huatuogpt} utilized a specialized medical validator.




\paragraph{ Preference Learning}

Preference learning is the dominant paradigm for aligning LLMs with human intent and values, as well as for enhancing complex capabilities such as reasoning. Schulman et al. \cite{schulman2017proximal} stood as a prominent example of this approach.
Preference learning trains a reward model not by defining absolute reward scores directly, but rather by comparing different model-generated outputs, typically on a pairwise basis, in order to reflect human preferences or other predefined criteria such as higher quality outputs or more rigorous reasoning processes.
This learned reward model subsequently serves as a reward signal to guide the optimization of the LLM's policy. PPO \cite{schulman2017proximal} is a commonly employed policy optimization algorithm within RLHF, with research such as \cite{zheng2023secrets} focusing on enhancing its stability and effectiveness through improved reward modeling and alignment with human values (e.g., usefulness, honesty, harmlessness). DeepSeekMath \cite{shao2024deepseekmath}, for instance, proposed a variant named Group Relative Policy Optimization (GRPO) to optimize memory usage.

Recently, DPO \cite{rafailov2024direct} has garnered significant attention as a simpler and potentially more stable alternative to the explicit reward modeling step inherent in RLHF. DPO optimizes the policy model directly from preference data and has been employed in training models such as Phi-4 \cite{abdin2024phi} and BOLT \cite{pang2025bolt}. Similarly, the Llama-berry \cite{zhang2024llama} framework utilizes a Pairwise Preference Reward Model (PPRM) for the global evaluation of distinct reasoning paths.
Furthermore, Self-Play and Self-Training mechanisms are also utilized to learn preferences, aiming to reduce reliance on external human annotations. For example, Self-Play Fine-tuning (SPIN) \cite{chen2024self} enables models to enhance their capabilities through adversarial interactions with their own instances, while Reinforced Self-Training (ReST) \cite{gulcehre2023reinforced} combines offline RL and self-training for iterative policy optimization. Many RL research efforts aimed at enhancing reasoning capabilities, including DeepSeek-R1 \cite{guo2025deepseek}, Kimi k1.5 \cite{team2025kimi}, and OpenRFT \cite{zhang2024openrft}, often implicitly optimize based on some form of preference (e.g., process quality, outcome correctness, or a combination thereof), even when this is not explicitly stated in their methodology.

\subsection{Self-Evolution}
\label{sect: self-evolution}
Self-evolution in reasoning models describes the process whereby a model utilizes its intrinsic capabilities or interacts with its environment, potentially including self-generated data or feedback, to progressively improve its performance on reasoning, problem-solving, or specific tasks. 
Crucially, this enhancement occurs without requiring continuous, intensive human supervision. 
Such a paradigm aims to lessen the dependence on large-scale, high-quality human-annotated datasets. 
Furthermore, it explores the potential for models to overcome the limitations inherent in their initial training data, thereby enabling continuous improvements in their capabilities. 
Typically, self-evolution proceeds through one or more iterative cycles. Within each cycle, the model generates outputs, evaluates its own performance, and subsequently refines its internal representations or knowledge base. 
This iterative refinement process leads to superior performance in subsequent cycles. 
The core mechanisms facilitating self-evolution include Self-evaluation and Feedback, Verification and Correction, Reinforcement Learning and Self-training.

\subsubsection{Self-Evaluation and Feedback}

Self-evaluation and feedback are crucial components of the Self-evolutionary process. This capability refers to the model's ability to evaluate the quality of its own generated outputs—such as reasoning steps, final answers, and confidence scores—and subsequently utilize this evaluation as a feedback signal to guide future actions. 


A key aspect of this is self-critique and feedback generation, where the model acts as a critic, analyzing its own output and providing suggestions for improvement. 
For instance, Madaan et al. \cite{zhang2024llama} employed the same LLM to evaluate its initial output and provide feedback, thereby guiding subsequent refinement iterations. 
Similarly, Tang et al. \cite{tang2025enabling} utilized a contrastive critique approach, enabling the model to analyze reference solutions, critique student-generated solutions, and produce structured critiques with corrective suggestions. 
Furthermore, frameworks like CRITIC \cite{gou2023critic} and AutoMathCritique \cite{xi2024enhancing} also leverage model-generated critiques, integrating them with feedback from external tools for verification purposes.

\subsubsection{Reinforcement Learning and Self-training}

Reinforcement Learning serves as a foundational training paradigm for enabling autonomous model evolution, leveraging self-generated data or feedback signals to drive the learning process.
reasoning LLM can enhance their capabilities through self-driven strategies, notably Self-Training and Self-Play.

Self-Training typically adheres to an iterative ``Generate-Filter-Learn" cycle, aiming to optimize the model using autonomously produced data:
1) \textbf{Generate}: Initially, the model proactively generates a substantial volume of candidate solutions or reasoning trajectories for specific tasks, such as mathematical problem-solving or code generation. Methodologies like STaR \cite{zelikman2024star}, rStar-Math \cite{guan2025rstar}, V-STaR \cite{hosseini2024v}, Quiet-STaR \cite{zelikman2024quiet}, ReST \cite{gulcehre2023reinforced}, $ReST^{EM}$ \cite{singh2024beyond}, and B-Star \cite{zeng2024b} incorporate this generative step.
2) \textbf{Filter}: Subsequently, a validation mechanism is employed to select high-quality or correct samples. This mechanism can range from relatively simple procedures, like verifying the final answer's correctness \cite{zelikman2024star} or executing unit tests, to more sophisticated techniques. Examples include scoring samples using a pre-trained reward model \cite{hosseini2024v,singh2024beyond} or selecting superior reasoning paths based on MCTS evaluation values \cite{guan2025rstar}.
3) \textbf{Learn}: Finally, the model undergoes fine-tuning, commonly through Supervised Fine-Tuning (SFT), on the curated high-quality dataset. This stage facilitates the assimilation of generated knowledge and enhances reasoning proficiency. 
To further bolster learning efficiency and efficacy, researchers have explored alternative learning strategies. For instance, RISE \cite{qu2025recursive} utilizes self-distillation to learn from generated improved data, whereas $ReST^{EM}$ \cite{singh2024beyond} advocates for training from the base model rather than iterative fine-tuning to mitigate potential error accumulation.

In contrast to the reliance on explicit filtering in self-training, Self-Play introduces an adversarial learning mechanism. Exemplified by the SPIN \cite{chen2024self}, the model is tasked not only with generating its own training data but also with learning to discriminate between these self-generated data and high-quality, human-annotated data. This competitive dynamic incentivizes the model to produce outputs that are increasingly indistinguishable from the reference data, thereby fostering iterative capability enhancement through the adversarial process.

Common to both self-training and self-play, Iteration is the fundamental engine driving continuous model evolution. Numerous aforementioned methods \cite{guan2025rstar,qu2025recursive,hosseini2024v,zelikman2024star,chen2024self,gulcehre2023reinforced,singh2024beyond,zeng2024b} explicitly or implicitly employ iterative execution. Within each iteration, the model leverages its current capabilities to generate data, undergoes evaluation, filtering, or adversarial comparison, and enhances its proficiency via the learning stage. This augmented capability subsequently enables the model to generate higher-quality data or make more optimal decisions in the subsequent round, establishing a positive feedback loop that progressively converges towards superior reasoning performance. Quiet-STaR \cite{zelikman2024quiet} further extends STaR's iterative learning concept, enabling the model to induce and learn reasoning rationales from arbitrary text, showcasing the potential of iterative improvement frameworks.



%% file: tabs/summary_of_REINFORCED_LEARNING.tex
\begin{table*}[t!]
\centering
\small
\caption{Summary of reinforced learning. TD and MTS mean Training data acquisition and Multi-Stage Training Strategies. RB means Rule-Based Reward.  }
\label{table: REINFORCED LEARNING}
\setlength{\tabcolsep}{0.3mm}{
\begin{tabular}{>{\raggedright\arraybackslash}m{0.15\textwidth} lll|cccccc}
\hline
Methods  & Policy & Reward & Self-evolution & MATH & AIME 2024 & MATH500 & MMLU & GSM8K & GPQA\\ 
\hline
DeepSeek-R1 \cite{guo2025deepseek} & MTS & RB & Self-training & & 79.8& 97.3& 90.8& - & 71.5    \\
T1 \cite{hou2025advancing} & MTS & Hybrid  & Self-training  &  -  &  50.6  &  92.4  & - & - & 56.1\\
BOLT \cite{pang2025bolt}& TD \& MTS  & ORM \& RB & Self-training  & - &  -  & 74.2 & - & - & - \\
AutoMath Critique \cite{xi2024enhancing} & MTS  & PRM & Feedback & 65.6 & - & - & - & - & 90.3 \\
STILL-2 \cite{Slow_Thinking_with_LLMs_3} & TD \& MTS  & ORM & Self-training & - & 46.7 & - & - & - & 51.0 \\
MATH-SHEPHERD \cite{wang2024math}& MTS  & PRM & Self-training  & - &  -  & 43.5 &   - & 89.1 & -  \\
MiniLLM \cite{gu2024minillm}& MTS & PRM & -  & - &  -  &  -  &  -  & - & -  \\
Phi-4  \cite{abdin2024phi}& MTS \& TD  & ORM &  Feedback & 80.4 & -   &  -  &  84.8  & - & 56.1  \\
RISE \cite{qu2025recursive}& TD \& MTS & ORM & Self-training  & 18.4 & - & - & - & 59.2 & -  \\
ReST \cite{gulcehre2023reinforced}& MTS \& TD & ORM &  Self-training  & - & - & - & - & - & -  \\
SPIN \cite{chen2024self}& MTS & ORM & Feedback  & - & - & - & 60 & 39.5 & - \\
SCoRe \cite{kumar2024training}&MTS&ORM&Self-training& 64.4 & - & - & - & - & - \\
OpenRFT \cite{zhang2024openrft} & TD & Hybrid & Self-training & - & - & - & - & - & - \\
OpenR \cite{wang2024openrt} & TD & PRM & Self-training & - & - & - & - & - & -\\

\hline
\end{tabular}
}
\end{table*}

%% file: secs/06SlowThinking.tex
\section{Slow Thinking}
\label{sect: slow thinking}
Slow Thinking-enhanced Long CoT integrates distilled training data, extended context processing, self-corrective reflection/backtracking, and implicit reasoning to emulate deliberate, human-like problem-solving in complex multi-step tasks (Table \ref{table:slow_thinking}). 

\subsection{Long CoT}
\label{sect: long cot}
The capacity to generate extended Chain-of-Thought sequences, often termed Long CoT, is fundamental for enabling large language models to tackle complex reasoning tasks requiring multi-step deliberation. Achieving robust Long CoT capabilities necessitates addressing challenges related to generation, context length, efficiency, and reliability. Consequently, recent research has explored several key strategies. These include techniques for data distillation to transfer sophisticated reasoning patterns, methods for extending context windows and improving reasoning over long sequences, approaches to foster implicit reasoning for greater efficiency, and mechanisms for reflection and backtracking to enhance accuracy and robustness. This section reviews recent advancements across these critical dimensions, detailing representative methodologies and their contributions to advancing Long CoT performance.

\subsubsection{Data Distillation}
\input{tabs/Summary_of_slow_thinking}
Data distillation via SFT has emerged as a prominent technique for imparting complex reasoning abilities, particularly long CoT capabilities, from large teacher models to smaller student models. For instance, Wu et al. \cite{wu2024comparativestudyreasoningpatterns} demonstrated that SFT can effectively transfer a teacher model’s explicit reasoning chains to a student model, enabling the latter to internalize both explicit and implicit reasoning patterns. Building on this, Guo et al. \cite{guo2025deepseek} employed an SFT-only strategy, foregoing a reinforcement learning stage, by designing specific objectives and loss functions tailored to different reasoning levels. This approach allows smaller models to acquire sophisticated reasoning skills from their larger counterparts. Illustratively, consider a teacher model generating a detailed, step-by-step reasoning trace for a complex task; through SFT using such data, the student model learns to replicate these intermediate steps and generalize the underlying reasoning process to novel problems. Further refining the granularity of distillation, Xiang et al. \cite{xiang2024atomthinkslowthinkingframework} introduced the AtomThink framework. This framework utilizes a CoT annotation engine coupled with step-level masking to decompose reasoning processes into atomic steps, training the student model to generate each intermediate step accurately for complex reasoning tasks. Addressing the efficiency of generated reasoning, Ma et al. \cite{ma2025cot} proposed CoT-Valve, a method that identifies parameter space directions to control the verbosity of the generated CoT. This technique facilitates the distillation of not only the teacher's explicit reasoning logic but also an efficient implicit reasoning process into the student model. Finally, Yu et al. \cite{yu2024distilling21} explored distilling System 2 capabilities (representing slower, deliberate reasoning) into student models exhibiting System 1-like efficiency (i.e., faster inference and lower resource usage). By training teacher models on explicit CoT and subsequently distilling this knowledge, their method allows student models to achieve strong performance in long-chain reasoning while significantly reducing computational overhead.

\subsubsection{Long-Context Extension \& Improvement}
Recent advancements have significantly extended the context processing capabilities and reasoning proficiency of large language models. The Kimi k1.5 model \cite{team2025kimi}, for example, features an extended context window of 128K tokens. This capability is supported by optimized attention mechanisms, such as Shifted Sparse Attention, which enable the model to process extensive contextual history while balancing computational efficiency with the scope of attention during inference. Addressing reasoning over long contexts, Zhao et al. \cite{zhao2024marco} introduced the Marco-o1 framework. This framework employs MCTS to generate synthetic long-chain CoT data, thereby improving the model's reasoning performance on tasks requiring extended context understanding. Similarly, Yang et al. \cite{yang2023iterative} explored context expansion through a "Stepwise Internalization" process. This method generates synthetic training data (e.g., for multi-digit multiplication tasks) designed to progressively internalize intermediate reasoning steps, enhancing both the model's reasoning capacity and its effectiveness in processing long documents. Furthermore, the Phi-4 technical report by Abdin et al. \cite{abdin2024phi} demonstrated that extending the context length from 4K to 16K tokens mid-training significantly improves model stability and accuracy on long-chain reasoning tasks.

\subsubsection{Implicit Reasoning}

Implicit reasoning refers to the capability of models to perform structured, step-by-step problem-solving internally, without necessarily verbalizing every intermediate computation or deduction. A common approach involves employing special tokens or designated markers during training or inference to encourage an internal simulation of CoT processes. For instance, Kimi k1.5 \cite{team2025kimi} employs markers such as <think> and </think> to structure its internal reasoning process, guiding it towards multi-step solutions. Alternatively, Deng et al. \cite{deng2024explicit} proposed a method to foster implicit CoT reasoning by gradually removing intermediate reasoning steps (tokens) during training. This process compels the model to internalize the reasoning pathway rather than explicitly generating each step. Furthermore, Liang et al. \cite{liang2023encouraging} utilized a multi-agent debate (MAD) framework designed to stimulate divergent thinking. The framework facilitates debates where agents present opposing arguments, compelling the model towards deeper logical analysis and consequently improving its capacity for implicit reasoning.

\subsubsection{Reflection and Backtracking}

Reflection and backtracking mechanisms facilitate a model's ability to monitor internal reasoning processes, detect errors, and dynamically adjust its reasoning trajectories. Numerous studies have incorporated techniques for overseeing intermediate reasoning steps. For instance, Guo et al. \cite{guo2025deepseek} proposed a Self-Refinement mode wherein the model continuously evaluates and, if necessary, corrects intermediate outputs via recursive checks. Similarly, Xu et al. \cite{xu2025redstar} presented the RedStar framework, which employs a "step-by-step verification" process to identify and flag errors within the generated reasoning chain, although its reported efficacy can be context-dependent. Furthermore, methods developed by Yu et al. \cite{yu2024distilling21} and Qi et al. \cite{qi2024interactive} specifically emphasize backtracking capabilities and associated techniques like gradient stabilization. Upon detecting a discrepancy during a reasoning step, the model can re-evaluate prior assumptions and adjust subsequent steps accordingly. For example, in a complex mathematical derivation, identification of an error at a specific computational step allows the model to backtrack, recalculate, and revise subsequent steps. Moreover, Min et al. \cite{min2024imitateexploreselfimprovereproduction} described a self-improvement paradigm where the model iteratively generates high-quality reasoning demonstrations, which are then incorporated into its training data. This continuous learning mechanism enables the model to progressively refine its reasoning strategies. Finally, Gan et al. \cite{gan2025rethinkingexternalslowthinkingsnowball} employed tree search algorithms, such as MCTS, to explore multiple potential reasoning paths. Reward models were subsequently utilized to select the optimal path, thereby effectively 
reflection and backtracking into the model's decision-making framework.

\subsection{Hierarchical Reasoning}
\label{sect: hierarchical reasoning frameworks}

Hierarchical reasoning frameworks have emerged as a crucial strategy in large model inference, specifically designed to overcome the limitations of monolithic models in tackling complex, multi-step problems. By imposing modularity—either through explicit structures, agentic collaboration, dynamic processes, or latent representations—these frameworks aim for more controllable, interpretable, and robust reasoning. This section surveys recent advancements, highlighting the specific challenges they address and the novel mechanisms they introduce.


\paragraph{Explicit Structures}  
Explicit structuring techniques seek improved control. ReasonFlux \cite{2025arXiv250206772Y} counters static reasoning paths by introducing dynamic pathfinding via Hierarchical Reinforcement Learning (HRL). Concurrently, Li et al. \cite{li2025search} utilized a bi-layered agentic Retrieval-Augmented Generation (RAG) and refinement architecture specifically designed to curtail error cascades through controlled, on-demand knowledge integration.


\paragraph{Agentic Systems} 
Agentic collaboration and tool integration significantly augment model capabilities. MALT \cite{motwani2025maltimprovingreasoningmultiagent} automates the optimization of distinct agent roles (generate, validate, refine). Complementary frameworks like OctoTools \cite{lu2025octotoolsagenticframeworkextensible} innovate through standardized tool encapsulation, while Agentic Reasoning \cite{wu2025agenticreasoningreasoningllms} synergizes internal knowledge structuring (e.g., mind maps) with external tool access for complex research domains.


\paragraph{Dynamic Control Mechanism}   
Addressing context-sensitivity and resource constraints, dynamic mechanisms offer enhanced flexibility. MixLLM \cite{wang2025mixllmdynamicroutingmixed} implements hierarchical meta-decision making for cost-aware dynamic query routing. AdaptiveStep \cite{liu2025adaptivestepautomaticallydividingreasoning}, conversely, introduces dynamic segmentation of reasoning processes based on model confidence, optimizing computational resource allocation.

\paragraph{Latent Space Manipulations} 
Innovations increasingly target the model's internal processes and representations. \cite{hao2024traininglargelanguagemodels}
Strategies include iterative refinement for enhanced In-Context Learning (Yang et al. \cite{yang2023iterative}, the introduction of explicit latent thought vectors for modular control \cite{kong2025scalable}, adversarial training frameworks for intrinsic permutation robustness \cite{chen2025pearlpermutationresilientllms}, and classifier-guided exploration of latent reasoning pathways \cite{wang2025thoughtprobeclassifierguidedthoughtspace}.

Collectively, these developments demonstrate a significant trend towards more sophisticated and diverse hierarchical architectures. By leveraging modularity across explicit structures, agentic systems, latent space manipulations, and dynamic control mechanisms, these frameworks substantially advance the capacity of large models for robust and complex reasoning.

\subsection{Hybrid Thinking}
\label{sect: Hybrid Thinking model}
Hybrid Thinking Mode (HTM) frameworks, inspired by dual-process cognitive theories \cite{kahneman2011thinking, booch2021thinking}, enhance large model inference by integrating rapid, intuitive processing (System 1) with deliberate, logical reasoning (System 2). This paradigm aims to overcome the limitations of single-mode processing, enabling adaptive reasoning strategies tailored to task complexity.

\paragraph{ Guided Search}
Key HTM implementations focus on orchestrating the interplay between fast and slow processes, often leveraging explicit control or search algorithms. 
Frameworks like HDFlow \cite{yao2024hdflow} dynamically combine direct CoT reasoning with complex workflow decomposition, while architectures such as Dualformer \cite{su2024dualformer} embed this duality structurally. Search and planning algorithms are frequently adapted: HaluSearch \cite{zhang2024fast} uses MCTS for guided slow generation to mitigate hallucinations; Q* \cite{deng2024q} employs a Q-value model for heuristic guidance of LLM generation; Mulberry \cite{yao2024mulberryempoweringmllmo1like} enhances MCTS with collective MLLM knowledge for reflection; and Flow-of-Options (FoO) \cite{nair2025flowofoptionsdiversifiedimprovedllm} systematically explores diverse reasoning paths.

\paragraph{Adaptive Control}
Another prevalent mechanism within HTM is adaptive control based on task or model state, which allows dynamic adjustments to the reasoning strategy.
DAST \cite{shen2025dastdifficultyadaptiveslowthinkinglarge} adjusts CoT length based on estimated problem difficulty; Entro-duction \cite{zhang2025entropybasedexplorationconductionmultistep} modulates search depth using model output entropy; and SIFT \cite{zeng2025siftgroundingllmreasoning} triggers slower refinement based on prediction discrepancies arising from factual ``Stickers".

\paragraph{Specialized Architectures}
HTM principles also manifest in specialized architectures and collaborative model setups, demonstrating structural ways to embody the dual-process approach. 
Examples include agent systems with distinct ``Talker" (fast) and ``Reasoner" (slow) roles \cite{christakopoulou2024agents}, collaborations between large (slow) and small (fast) models like FS-GEN \cite{zhang2024fast}, skill-based Mixture-of-Experts routing (SYMBOLIC-MoE \cite{chen2025symbolicmixtureofexpertsadaptiveskillbased}), and neuro-symbolic tools like Lemmanaid \cite{alhessi2025lemmanaidneurosymboliclemmaconjecturing} combining fast neural generation with slow symbolic validation.

\paragraph{Tailored Training}
Furthermore, HTM concepts increasingly influence model training strategies and internal components, showing the paradigm's impact beyond high-level control. 
Techniques include aligning autoregressive models with iterative processors (RELAY \cite{yu2025enhancingautoregressivechainofthoughtloopaligned}), distilling mixed-complexity reasoning paths (Mix Distillation \cite{li2025smallmodelsstrugglelearn}), dynamically gating attention (MoBA \cite{lu2025mobamixtureblockattention}), and balancing exploration-exploitation during self-training (B-STaR \cite{zeng2025bstarmonitoringbalancingexploration}). The paradigm's applicability extends to diverse domains, including multimodal visual reasoning (FAST \cite{sun2024visual}, Hu et al. \cite{hu2023tree}), mathematical problem-solving (MathFusion \cite{pei2025mathfusionenhancingmathematicproblemsolving}), and adaptable rule-following (Meta-RFFT \cite{hu2025traininglargelanguagemodels}).

In essence, HTM frameworks achieve enhanced reasoning by dynamically integrating rapid intuition and deliberate logic. This is realized through diverse mechanisms, including guided search, adaptive control, specialized architectures, and tailored training, collectively improving the efficiency, robustness, and adaptability of large models on complex tasks.

%% file: tabs/Summary_of_slow_thinking.tex
\begin{table*}[t!]
\centering
\scriptsize
\caption{Summary of Slow Thinking Performance. DD, CE, IR, and RB mean Data Distillation, Long-Context Extension \& Improvement, Implicit Reasoning, Reflection and Backtracking. ES, AS, LSM, and DCM mean Explicit Structures, Agentic Systems, Latent Space Manipulations, and Dynamic Control Mechanisms. GS, AC, SA, and TT mean Guided Search, Adaptive Control, Specialized Architectures, and Tailored Training.}
\label{table:slow_thinking}
\setlength{\tabcolsep}{0.6mm}{
\begin{tabular}{llll|cccccccc}
\hline
Paper & Long CoT & Hierarchical & Hybrid & AIME & MATH500 & MMLU & GSM8K & MGSM & LiveCodeBench & HumanEval \\ 
\hline
DeepSeek-R1 \cite{guo2025deepseek} & IR \& DD \& RB & ES & TT & 79.8 & 97.3 & 90.8 & - & - & 65.9 & -\\
phi-4 \cite{abdin2024phi} & CE & AS & TT & - & - & 84.8 & - & 80.6 & - & 82.6 \\
Marco-o1 \cite{zhao2411marco} & DD \& RB & ES \& DCM & GS \& AC \& TT & - & - & - & - & 88.4 & - & -\\
Kimik1.5 \cite{team2025kimi} & CE \& IR \& RB & - & TT & 77.5 & 96.2 & 87.4 & - & - & 62.5 & 81.5 \\
CoT-Valve \cite{ma2025cot} & CE \& DD & LSM \& DCM  & AC\&TT & - & - & - & 94.9 & - & - & - \\
MoBA \cite{MoonshotMoBA} & CE \& DD & DCM & AC \& SA \& TT & - & - & 49.0 & 72.8 & - & - & 69.5 \\
Heima \cite{shen2025efficientreasoninghiddenthinking} & IR & LSM & TT & - & - & - & - & -  & - & - \\

PROMPTCOT \cite{zhao2025promptcotsynthesizingolympiadlevelproblems} & DD & ES & GS & - & 93.0 & - & 92.6 & - & - & - \\

RealSafe-R1
 \cite{zhang2025realsafer1safetyaligneddeepseekr1compromising} & DD & ES & TT & - & 95.7 & - & - & - & - & - \\

O1-Pruner \cite{luo2025o1prunerlengthharmonizingfinetuningo1like} & RB & DCM & GS & - &  - & 96.5 & - & - & - & - \\
\hline
\end{tabular}
}
\end{table*}

%% file: secs/08Challenges.tex
\section{Challenges and Further Directions}
\label{Challenges and Further Directions}

\paragraph{The Balance between Fast and Slow Thinking}

Achieving a balance between fast and slow thinking in LLMs remains a significant challenge. While fast thinking allows models to generate quick, intuitive responses based on learned patterns, slow thinking enables deliberate, step-by-step reasoning, which is essential for solving complex problems. However, integrating these two modes effectively is non-trivial. Though some studies try to combine fast and slow thinking (e.g., Claude 3.7\footnote{https://www.anthropic.com/news/claude-3-7-sonnet} and Qwen 3\footnote{https://github.com/QwenLM/Qwen3}), current LLMs predominantly operate in a fast-thinking mode, relying on pre-trained knowledge and pattern recognition. To incorporate slow thinking, models must be trained to produce extended chains of reasoning while maintaining coherence and accuracy. Future research should focus on designing hybrid architectures that dynamically switch between fast and slow thinking based on task requirements, ensuring both efficiency and depth in reasoning.

\paragraph{Multi-modal Reasoning Large Language Model}

Extending slow-thinking capabilities to multi-modal reasoning represents another promising direction. Real-world problems often involve multiple modalities, such as text, images, audio, and video. For instance, reasoning about a scientific experiment may require interpreting textual descriptions, visual diagrams, and numerical data simultaneously \cite{zhang2023multimodal}. Current LLMs are primarily text-based, limiting their ability to handle such multi-modal tasks. Developing multi-modal reasoning models that can integrate information from diverse sources while engaging in slow, deliberate reasoning will significantly enhance their applicability. Challenges include aligning representations across modalities, ensuring consistency in reasoning, and scaling models to handle the increased complexity of multi-modal inputs.

\paragraph{Reinforcement Learning Stability and Reward Design}

RL-based fine-tuning, such as RLHF or RLAIF, is crucial for improving reasoning capabilities in LLMs. However, these methods often suffer from training instability and reward hacking \cite{miao2024inform}, where models exploit loopholes in the reward function to achieve high scores without genuinely improving reasoning quality.
Designing robust reward models that align with reasoning quality rather than superficial patterns is a non-trivial task. For example, rewarding models solely based on correct answers may lead to overfitting to specific benchmarks while neglecting the reasoning process itself. In subjective domains, scaling reinforcement learning with increasingly complex inputs can be hindered if the reward model becomes a bottleneck in providing appropriate reward signals. Ensuring that the reward model keeps up with the complexity of tasks is crucial for effective learning. Future work should explore novel reward design strategies, such as incorporating intermediate reasoning steps into the reward function or leveraging human-in-the-loop feedback to refine reward signals dynamically.

\paragraph{Generalization vs. Over-Optimization}

One of the risks of training slow-thinking models is overfitting to specific reasoning benchmarks, such as GSM8K (grade-school math problems) or MATH (advanced mathematical problems). While these benchmarks provide valuable training data, they may not fully capture the diversity and complexity of real-world problem-solving scenarios. Models that perform well on benchmarks may struggle when faced with unfamiliar tasks or domains. Ensuring cross-domain robustness is essential for practical deployment. Future research should focus on developing techniques to improve generalization, such as augmenting training data with diverse problem types, introducing domain-specific constraints, and evaluating models on out-of-distribution tasks. Additionally, exploring meta-learning approaches that enable models to adapt quickly to new domains could further enhance their versatility.

\paragraph{Self-Improving RL Frameworks}

Exploring self-improving reinforcement learning frameworks, such as meta-reinforcement learning or iterative self-training, represents an exciting direction for advancing slow-thinking models. In these frameworks, models learn to refine their own reasoning policies over time by iteratively generating new training data, evaluating their performance, and updating their strategies. For example, a model could generate potential solutions to a problem, simulate their outcomes, and use the results to improve its reasoning process. This approach mimics human learning, where individuals reflect on past experiences to enhance future performance. Key challenges include ensuring stability during self-improvement, avoiding catastrophic forgetting, and scaling the framework to handle increasingly complex tasks. Successful implementation of self-improving RL frameworks could lead to models that continuously evolve and adapt, achieving higher levels of reasoning capability.

\paragraph{Human-in-the-Loop Refinement}

Incorporating human-in-the-loop refinement is another promising avenue for enhancing slow-thinking models. Human feedback can provide valuable insights into areas where models struggle, such as ambiguous reasoning steps or incorrect assumptions. Interactive feedback mechanisms, such as debate systems or iterative correction workflows, allow humans to guide models toward better reasoning strategies. For instance, in a debate system, multiple models could present competing arguments, and human judges could evaluate their reasoning quality. Similarly, in an iterative correction workflow, humans could identify errors in a model's reasoning trace and suggest corrections, which the model then incorporates into its training process. Leveraging human expertise in this way can help refine slow-thinking models in real-world scenarios, improving their reliability and robustness.

\paragraph{Application of Other Domains}

Extending slow-thinking models to other domains, such as robotics, recommendation systems, and healthcare, offers immense potential for impact. In robotics, slow-thinking capabilities could enable robots to plan complex actions, reason about uncertainties, and adapt to dynamic environments. For example, a robot tasked with assembling furniture could use slow thinking to break down the task into manageable steps, anticipate potential obstacles, and adjust its strategy accordingly. In recommendation systems, slow-thinking models could analyze user preferences more deeply, considering long-term trends and contextual factors to provide personalized suggestions. In healthcare, slow-thinking models could assist doctors in diagnosing diseases, interpreting medical data, and designing treatment plans by engaging in thorough, evidence-based reasoning. Each domain presents unique challenges, such as integrating domain-specific knowledge, handling real-time constraints, and ensuring safety and reliability. Addressing these challenges will require interdisciplinary collaboration and domain-specific adaptations of slow-thinking models.